\documentclass{article}

    \usepackage[preprint]{neurips_2023}

\usepackage[utf8]{inputenc} 
\usepackage[T1]{fontenc}    
\usepackage{hyperref}       
\hypersetup{colorlinks,linkcolor=red,
            anchorcolor=blue,
            citecolor=green}
\usepackage{url}            
\usepackage{booktabs}       
\usepackage{amsfonts}       
\usepackage{nicefrac}       
\usepackage{microtype}      
\usepackage{xcolor}         
\usepackage{multirow}
\usepackage{graphicx}
\usepackage{wrapfig}
\usepackage{mathtools}
\usepackage{makecell} 
\usepackage{lipsum}
\usepackage{enumitem}
\usepackage{subfigure}
\usepackage{bbding}
\usepackage[misc]{ifsym}
\usepackage{appendix}
\usepackage{natbib}
\setcitestyle{numbers,square}

\title{Benchmarking Test-Time Adaptation against Distribution Shifts in Image Classification}

\author{Yongcan Yu$^{1,3}$, \ Lijun Sheng$^{1,2}$, \ Ran He$^{1,3}$, \ Jian Liang$^{1,3}$ \textsuperscript{\Letter}\\
$^{1}$ CRIPAC \& MAIS, Institute of Automation, Chinese Academy of Sciences \\
$^{2}$ University of Science and Technology of China, 
$^{3}$ University of Chinese Academy of Sciences\\
{\tt\small yuyongcan0223@gmail.com, liangjian92@gmail.com}}

\begin{document}

\maketitle

\begin{abstract}
Test-time adaptation (TTA) is a technique aimed at enhancing the generalization performance of models by leveraging unlabeled samples solely during prediction. 
Given the need for robustness in neural network systems when faced with distribution shifts, numerous TTA methods have recently been proposed. 
However, evaluating these methods is often done under different settings, such as varying distribution shifts, backbones, and designing scenarios, leading to a lack of consistent and fair benchmarks to validate their effectiveness. 
To address this issue, we present a benchmark that systematically evaluates 13 prominent TTA methods and their variants on five widely used image classification datasets: CIFAR-10-C, CIFAR-100-C, ImageNet-C, DomainNet, and Office-Home. 
These methods encompass a wide range of adaptation scenarios (\textit{e.g.} online adaptation v.s. offline adaptation, instance adaptation v.s. batch adaptation v.s. domain adaptation). 
Furthermore, we explore the compatibility of different TTA methods with diverse network backbones. 
To implement this benchmark, we have developed a unified framework in PyTorch, which allows for consistent evaluation and comparison of the TTA methods across the different datasets and network architectures. 
By establishing this benchmark, we aim to provide researchers and practitioners with a reliable means of assessing and comparing the effectiveness of TTA methods in improving model robustness and generalization performance. 
Our code is available at \url{https://github.com/yuyongcan/Benchmark-TTA}.
\end{abstract}

\section{Introduction}

Nowadays, deep neural networks have demonstrated impressive performance in various domains, such as image classification \cite{resnet,xie2020self}, semantic segmentation \cite{strudel2021segmenter,xie2021segformer}, and natural language processing \cite{devlin2018bert,lewis2019bart}. 
Such high accuracies are always guaranteed under certain conditions, specifically when the training data and test data are sampled from the same distribution. 
Unfortunately, in the real world, there always exists a distribution shift \cite{hendrycks2021many, croce2021robustbench, kar20223d} between the training data and test data.
For instance, in the autonomous driving scenario, control systems developed by manufacturers may be trained using data collected in sunny weather conditions but are deployed in more complex and challenging environments such as rain and fog. In such cases, neural networks often experience a notable decrease in performance.

\begin{figure}[htbp]
\centering
\subfigure[TTBA]{\includegraphics[width=0.3\textwidth]{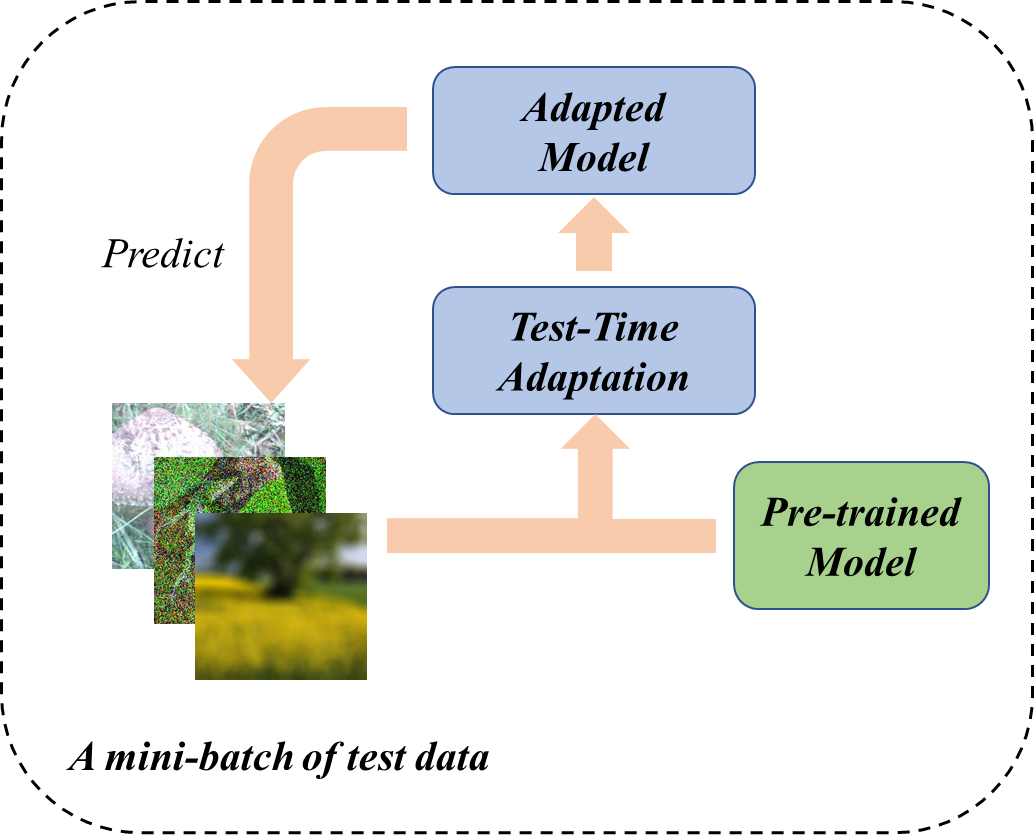}}
\subfigure[OTTA]{\includegraphics[width=0.3\textwidth]{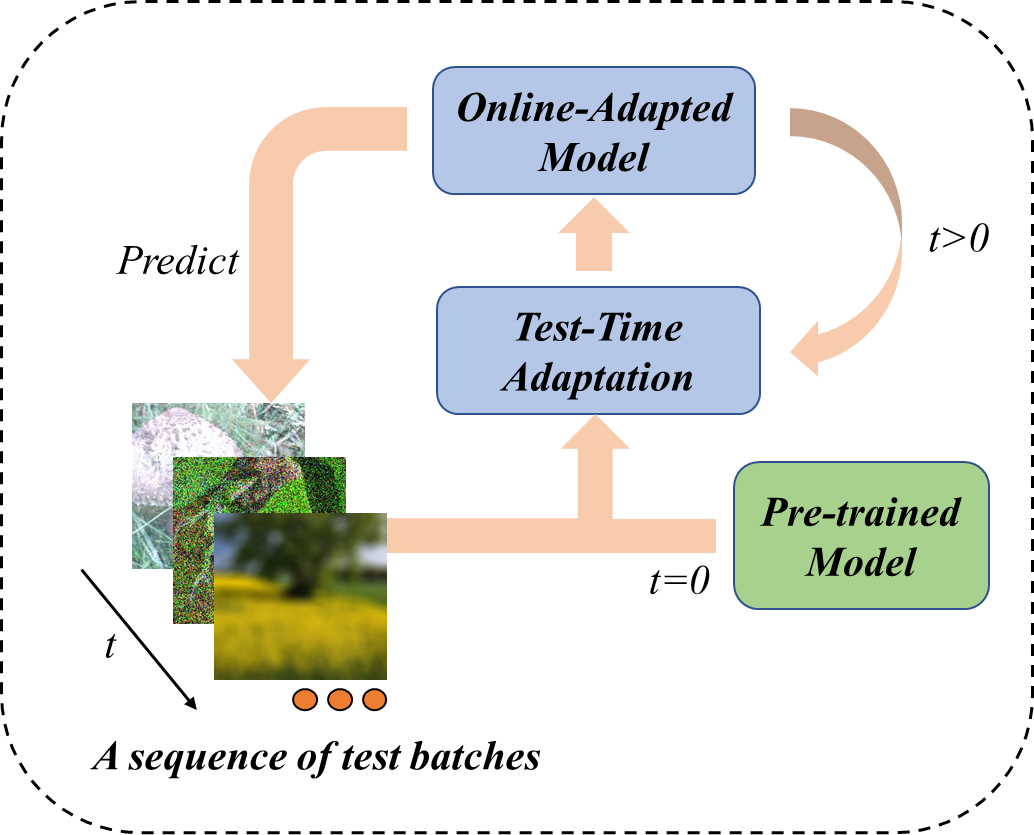}}
\subfigure[TTDA]{\includegraphics[width=0.31\textwidth]{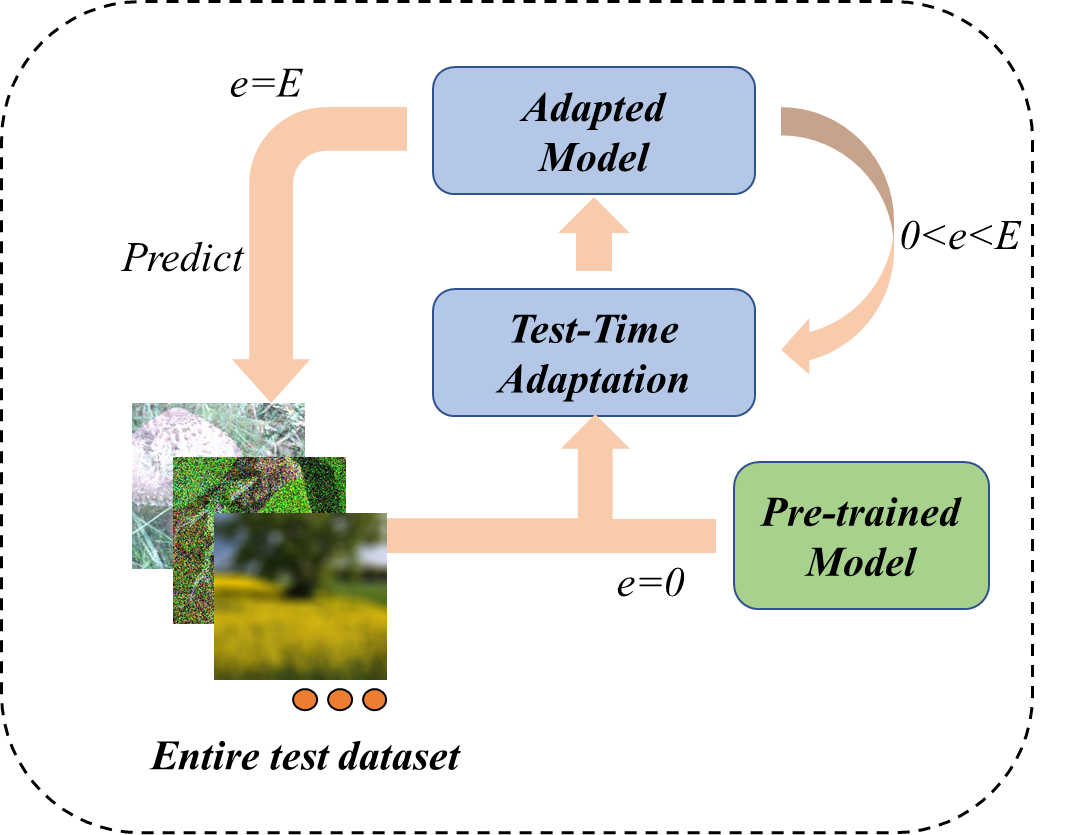}}
\caption{The illustration of three typical cases of TTA scenarios. $t, e, E$ denote the time-step, current and max epoch, respectively.}
\label{fig:workflow}
\vspace{-0.65cm}
\end{figure}

To mitigate the negative impact of distribution shifts, test-time adaptation (TTA) \cite{liang2023ttasurvey} has emerged as a significant approach. 
As shown in Figure \ref{fig:workflow}, unlike unsupervised domain adaptation (UDA) \cite{zhang2015deep, ghifary2016deep}, TTA does not require any labeled source data. 
Instead, it focuses on dynamically adjusting the behavior of a pre-trained model during inference based on the specific characteristics of the unlabeled test data. 
This allows the model to improve its performance when encountering variations, unseen examples, or changing conditions at test time.
Typically, there exist various TTA methods designed for different scenarios, such as test-time batch adaptation (TTBA) \cite{zhang2021adaptive,wang2022out}, online test-time adaptation (OTTA) \cite{gan2022decorate, chen2023improved}, and test-time domain adaptation (TTDA) \cite{liang2020we, shotplus, ding2022proxymix}, which assume test data come in single mini-batch, online batches, and domain fashion, respectively. 
Despite the strong correlation between these scenarios, previous works \cite{alfarra2023revisiting, marsden2022introducing} have not evaluated them within a unified framework. It is worth noting that there exists a concurrent work named TTAB \cite{zhao2023pitfalls} that discusses the impact of hyperparameters and various distribution shifts in TTA, but it only focuses on batch adaptation and online adaption methods.
In addition, most current TTA approaches are evaluated empirically using only corruption datasets (\textit{e.g.} CIFAR-10-C \cite{krizhevsky2009learning, hendrycks2019benchmarking}, CIFAR-100-C \cite{krizhevsky2009learning, hendrycks2019benchmarking}, and ImageNet-C \cite{deng2009imagenet, hendrycks2019benchmarking} for TTBA and OTTA) or natural-shift datasets (typically Office-Home \cite{venkateswara2017deep} and DomainNet \cite{peng2019moment, saito2019semi} for TTDA). 

In order to address the above two defects in evaluating TTA methods and facilitate general TTA research, we propose a benchmark that aims to provide a comprehensive and unbiased evaluation of current TTA methods.
For the sake of \textbf{comprehensiveness}, our benchmark delves into three key aspects: method, dataset, and model architecture. Regarding the method, we evaluate several TTA methods specifically designed for different scenarios \cite{liang2023ttasurvey}, including TTDA, OTTA, and TTBA. Additionally, we also explore variants of some of these methods. With respect to the dataset, our benchmark covers both corruption datasets and natural-shift datasets. Furthermore, to demonstrate the impact of model architecture, we apply both Convolutional Neural Networks (CNN) \cite{lecun1989backpropagation} and Vision-Transformer (ViT) \cite{dosovitskiy2020image} in our experiments.
In order to ensure \textbf{fairness} in our evaluation, we maintain consistency in crucial parameters within the TTA setting, including the batch size. Additionally, we adopt a unified hyperparameter search strategy to select appropriate hyperparameters for each TTA algorithm.

To conclude, our work has made significant contributions and provided valuable insights into the following aspects:
\begin{itemize}[leftmargin=0.4cm]
    \item \textbf{Benchmark Proposal}: We have introduced a comprehensive and equitable benchmark that evaluates 13 TTA methods  across five image classification datasets. Notably, our study is the first to investigate the effectiveness of various TTA methods on both corruption datasets and natural datasets. This holistic evaluation framework ensures a thorough examination of TTA methods' performance.
    \item \textbf{Performance Comparison}: By directly implementing OTTA methods for multiple epochs, OTTA methods achieve competitive accuracy results on corruption datasets when compared to TTDA methods. However, there remains a discernible gap between the performance of OTTA methods and TTDA methods on natural-shift datasets. This finding underscores the potential for further improvement in the domain generalization capabilities of OTTA methods.
    \item \textbf{Robustness of PredBN}: We have observed that PredBN \cite{nado2020evaluating} is both effective and efficient in handling corruption datasets. It demonstrates superior performance in mitigating the effects of corruption. However, PredBN and its derivative techniques do not yield the same level of performance on natural-shift datasets as they do on corruption datasets.

\end{itemize}


\section{A General Paradigm of Test-Time Adaptation}
\subsection{Problem Formulation} 
As a subfield of domain adaptation (DA), test-time adaptation (TTA) distinguishes itself from standard DA by not accessing the source data during the adaptation process. Let $f_{\theta_0}(x)$ be a model with parameter $\theta_0$ well-trained on the labeled source data $(\mathcal{X}_S, \mathcal{Y}_S)$. The objective of TTA is to adapt the source model $f_{\theta_0}(x)$ to unlabeled target data $\mathcal{X}_T$ without utilizing any source data, where $\mathcal{X}_T$ and $\mathcal{X}_S$ are sampled from distinct distributions.

\begin{table}[htbp]
\caption{The details of different paradigms.}
\label{tab:paradigms}
\begin{tabular}{lccc}
\toprule
 Paradigm & Source Data & Instant Prediction  & Online Update \\ \midrule
 Standard Domain Adaptation \cite{zhang2015deep, ghifary2016deep} & \Checkmark &\XSolidBrush &\XSolidBrush \\
 Test-Time Domain Adaptation \cite{liang2020we, yang2021exploiting} &\XSolidBrush &\XSolidBrush&\XSolidBrush \\
 Test-Time Batch Adaptation \cite{nado2020evaluating, zhang2022memo} &\XSolidBrush&\Checkmark&\XSolidBrush \\
 Online Test-Time Adaptation \cite{wang2020tent, niu2022efficient} &\XSolidBrush&\Checkmark&\Checkmark\\ \bottomrule
\end{tabular}
\end{table}

\subsection{Paradigms}
There are three primary paradigms for TTA: test-time domain adaptation (TTDA), test-time batch adaptation (TTBA), and online test-time adaptation (OTTA). In the following sections, we will present an introduction to each of these paradigms. The settings of different paradigms are shown in Table~\ref{tab:paradigms} and an illustration is shown in Figure \ref{fig:workflow}.

\textbf{Test-Time Domain Adaptation}. TTDA assumes that the entire target domain data can be observed simultaneously during adaptation. Consequently, TTDA often trains the source model on the target data for several epochs using self-supervised loss (\textit{e.g}. pseudo label \cite{lee2013pseudo}, entropy minimization \cite{grandvalet2004semi}, information maximization \cite{liang2020we}) and employ the features obtained from the intermediate layers of the adapted model to refine the pseudo labels for the entire target dataset \cite{liang2020we, yang2021exploiting}. Then, it returns the predictions $y_t = f_{\theta'}(x_t)$ for each $x_t \in \mathcal{X}_T$ collectively, where $\theta'$ represents the adapted parameters.
\textbf{Test-Time Batch Adaptation}. TTBA assumes that data is received by the system in mini-batches. For each incoming mini-batch data $x_t$ with a batch size $B\ge 1$, the model adjusts its parameters from $\theta_0$ to $\theta_t$ based on the characteristics of $x_t$. Subsequently, TTBA outputs the prediction $y_t = f_{\theta_t}(x_t)$. Due to its nature, the model can provide instant predictions as samples enter the system in batches. It is worth noting that TTBA can be referred to as Test-Time Instance Adaptation (TTIA) when $B=1$.

\textbf{Online Test-Time Adaptation}. In OTTA, test data arrives in a sequential fashion in the form of mini-batches. Similar to TTBA, OTTA is an instant prediction paradigm. At time step $t$, given an input sample batch with a batch size $B$, OTTA aims to update the model's parameters from $\theta_{t-1}$ to $\theta_t$ based on the input $x_t$. Subsequently, it classifies $x_t$ using $f_{\theta_t}(x_t)$. In OTTA, the model is incrementally updated online, and the knowledge acquired from previous samples influences the prediction for the current samples. Typically, the target domain of OTTA remains stationary. It's worth noting that when the distribution of the target domain undergoes dynamic changes, this paradigm can also be referred to as continual test-time adaptation (CTTA).

\section{Test-Time Adaptation Algorithms and Datasets for Evaluation}
\subsection{Test-Time Adaptation Algorithms}

Several studies \cite{wang2020tent, liang2023ttasurvey} have extensively investigated the concept of TTA in the fields of machine learning and computer vision. This section provides a comprehensive overview of the existing literature related to TTA, highlighting various approaches and techniques proposed to enhance model performance and generalization during the inference phase.

To address the scarcity of source data, Liang \emph{et al.} \cite{liang2020we} propose a method that narrows the gap between the source and target domain by utilizing information maximization and clustering-based pseudo-label refinement.  NRC \cite{yang2021exploiting} explores the intrinsic neighborhood structure by promoting class-consistency with neighborhood affinity. Similarly, Litrico \emph{et al.} \cite{litrico2023guiding} refine pseudo labels based on neighborhoods and present PLUE, a technique that minimizes the entropy of the prediction distribution. Drawing inspiration from MoCo \cite{he2020momentum}, Chen \emph{et al.} \cite{chen2022contrastive} introduce contrastive learning into TTA and utilize both pseudo-label loss and diversity loss for optimization.

However, the aforementioned research primarily focuses on prediction after a long-term adaptation, which renders it unsuitable for scenarios that required instant prediction. In response to this limitation, Nado \emph{et al.} \cite{nado2020evaluating} propose a simple yet effective solution called PredBN. PredBN replaces the means ($\mu_s$) and variances ($\sigma_s$) of batch normalization (BN) \cite{ioffe2015batch} Layers learned during the training stage with the statistics of test batches ($\mu_t, \sigma_t$). Nevertheless, PredBN performs optimally only when the test batch size is sufficiently large. To tackle this issue, PredBN+ \cite{schneider2020improving} interpolate the ($\mu_s, \sigma_s$) and ($\mu_t, \sigma_t$) with $\alpha$ and $1-\alpha$ to forward. This approach allows PredBN+ to balance the importance of source knowledge and the information provided by the test data. Another solution for limited batch size is MEMO \cite{zhang2022memo}. MEMO employs various data augmentation \cite{shorten2019survey} techniques for a single image, enabling the acquisition of more stable statistical parameters for BN. Furthermore, MEMO minimizes the entropy of the augmented samples' marginal distribution of predictions in a self-supervised manner. In contrast, LAME \cite{boudiaf2022parameter} proposes Laplacian adjusted maximum-likelihood estimation between the model's output and the sample features, without updating any model parameters, to enhance prediction accuracy.

Recognizing that valuable knowledge can be extracted from previous samples as test data arrives in a streaming fashion, Wang \emph{et al.} \cite{wang2020tent} introduce a novel paradigm called fully test-time adaptation, also known as OTTA. After that, numerous online update methods emerge. Tent \cite{wang2020tent} updates the BN affine parameters by minimizing the entropy of the model output. Different from Tent update BN parameters, T3A \cite{iwasawa2021test} is a back-propagation-free method that adjusts its classifier according to the prototypes during testing. Designing for the CTTA paradigm, CoTTA \cite{wang2022continual} adopts the teacher-student \cite{hinton2015distilling} framework and applies multiple augmentations to test samples. In addition, CoTTA introduces random weights restore mechanism to prevent the model from catastrophic forgetting. Both EATA \cite{niu2022efficient} and SAR \cite{niu2023towards} incorporate entropy-based sample selection to enhance efficiency and reduce error accumulation. In addition, EATA also employs weight regularization to ensure that the model does not deviate from the source model while SAR adopts Sharpness-Aware Minimization \cite{foret2020sharpness} to improve generalization.

For the evaluation of TTA, we choose various methods: SHOT \cite{liang2020we}, NRC \cite{yang2021exploiting}, AdaContrast \cite{chen2022contrastive}, PLUE \cite{litrico2023guiding}, PredBN \cite{nado2020evaluating}, PredBN+ \cite{schneider2020improving}, MEMO \cite{zhang2022memo}, LAME \cite{boudiaf2022parameter}, Tent \cite{wang2020tent}, CoTTA \cite{wang2022continual}, EATA \cite{niu2022efficient} and SAR \cite{niu2023towards}. The overview of these algorithms we chose is shown in Table \ref{tab:algorithms}.


\setlength{\tabcolsep}{5.0pt}
\begin{table}[htbp]
\centering
\caption{The description of TTA algorithms. Parameter Update represents the parameter that will be updated during adaptation. BN, CLS, and FE denote batch normalization layer, classifier, and feature extractor, respectively. Data Augmentation represents whether the algorithm applies data augmentation or not. And Contrastive implied whether the algorithm uses contrastive learning to update the model or not.}
\label{tab:algorithms}
\resizebox{0.8\textwidth}{!}{\begin{tabular}{clcccc}
\toprule
 Paradigm  & Algorithm & Parameter Update  & Augmentation  &Contrastive &Venue \\
 \midrule
 \multirow{4}{*}{TTBA}  & PredBN+ \cite{schneider2020improving} & - &\XSolidBrush&\XSolidBrush & NeurIPS'20  \\
 & MEMO \cite{zhang2022memo} & All &\Checkmark&\XSolidBrush & NeurIPS'22   \\
 & PredBN \cite{nado2020evaluating} & - &\XSolidBrush&\XSolidBrush& ICMLW'20    \\
 & LAME \cite{boudiaf2022parameter} & - &\XSolidBrush&\XSolidBrush & CVPR'22   \\ 
 \midrule
 \multirow{5}{*}{OTTA}& Tent \cite{wang2020tent} & BN  &\XSolidBrush&\XSolidBrush & ICLR'20 \\
 & T3A \cite{iwasawa2021test} & CLS &\XSolidBrush&\XSolidBrush & NeurIPS'21   \\
 & CoTTA \cite{wang2022continual} & All &\Checkmark &\XSolidBrush & CVPR'22  \\
 & EATA \cite{niu2022efficient} &  BN   &\XSolidBrush&\XSolidBrush & ICML'22   \\
 & SAR \cite{niu2023towards}  &  BN & \XSolidBrush&\XSolidBrush&  ICLR'23  \\ 
 \midrule
 \multirow{4}{*}{TTDA} & SHOT \cite{liang2020we} & FE &\XSolidBrush&\XSolidBrush & ICML'20  \\
 & NRC \cite{yang2021exploiting} &All  &\XSolidBrush & \Checkmark& NeurIPS'21 \\
 & AdaContrast \cite{chen2022contrastive} & All &\Checkmark  &\Checkmark & CVPR'22  \\
 & PLUE \cite{litrico2023guiding} & All &\XSolidBrush&\Checkmark& CVPR'23  \\ 
 \bottomrule
\end{tabular}}
\vspace{-0.5cm}
\end{table}
\subsection{Datasets for Evaluation}
In order to conduct a comprehensive assessment of various TTA methods, we have carefully selected five widely used datasets, which include corruption datasets as well as natural-shift datasets. 
In this section, we will provide a detailed introduction to these datasets, aiming to gain valuable insights into our benchmark.

\textbf{Corruption Datasets}. The corruption datasets we consider include CIFAR-10-C and CIFAR-100-C \cite{hendrycks2019benchmarking}, which are variants of the test sets of CIFAR-10 and CIFAR-100 \cite{krizhevsky2009learning}, respectively. These datasets introduce 15 artificial corruptions to the original test sets, each with five severity levels. The corruptions applied are as follows: \textit{gaussian noise, shot noise, impulse noise, defocus blur, glass blur, motion blur, zoom blur, snow, frost, fog, brightness, contrast, elastic transform, pixelate, jpeg compression}. Similarly, ImageNet-C \cite{ hendrycks2019benchmarking} is derived from the validation set of ImageNet \cite{deng2009imagenet}, where 19 different corruptions, also with five severity levels, are applied. In addition to the aforementioned 15 corruptions, ImageNet-C introduces an additional four: \textit{saturate, spatter, speckle noise, gaussian blur}. For the sake of convenience, we conduct our experiments on the shared 15 corruptions, each at severity level 5, in the corruption datasets.

\textbf{Natural-shift Datasets}. Apart from the corruption datasets, we also evaluate our methods on Office-Home \cite{venkateswara2017deep} and DomainNet \cite{peng2019moment, saito2019semi}. Office-Home is a widely used domain adaptation dataset, consisting of four domains and encompassing 65 everyday object categories. Each domain comprises a varying number of images: \textit{Artistic images} (2,427 images), \textit{Clip Art} (4,365 images), \textit{Product images} (4,439 images), and \textit{Real-World images} (4,357 images). DomainNet is a large-scale domain adaptation dataset containing six domains. For our evaluation, we utilize a subset of DomainNet known as DomainNet126 \cite{saito2019semi}, which includes 126 classes in DomainNet across four domains: \textit{Clipart (18,703 images), Painting (31,502 images), Real (70,358 images)}, and \textit{Sketch (24,582 images)}. Since most OTTA methods are only tested on the corruption dataset and not on the natural shift-based dataset, we introduce the Office-Home and DomainNet126 databases to provide more inspiration for research in the current field.

\section{Empirical Studies}
This section aims to demonstrate the methodology employed in conducting the benchmark, as well as to present the primary outcome derived from our benchmark analysis. Moreover, supplementary experiments have been carried out to further enhance our understanding of the TTA research.

\subsection{Implementation Details}
\textbf{Network Architecture}. For the experiments conducted on CIFAR-10-C and CIFAR-100-C datasets, we utilized the pre-trained models, namely WideResNet \cite{zagoruyko2016wide} and ResNeXt-29 \cite{xie2017aggregated}, respectively, which were made available through the RobustBench benchmark \cite{croce2021robustbench}. For the ImageNet-C experiments, we employed the pre-trained ResNet-50 \cite{resnet} model, using the \textit{IMAGENET1K\_V1} weights provided by the \textit{torchvision} library. Furthermore, for the ViT-based experiment on ImageNet-C, we utilized the pre-trained model \textit{vit\_base\_patch16\_224} \cite{dosovitskiy2020image} from the \textit{timm} library. For the Office-Home and DomainNet126 experiments, in terms of network structure and source model training, we follow \cite{shotplus}.

\textbf{Hyperparameter Search}. This paragraph outlines our approach to selecting hyperparameters. To ensure proper validation, we select the first task from each dataset as the validation task to employ a grid search. Specifically, we use \textit{original $\rightarrow$ gaussian noise, Artistic $\rightarrow$ Clip, Clipart $\rightarrow$ Painting} as the validation tasks for the corruption datasets, Office-Home, and DomainNet126, respectively. It is important to mention that we restrict the parameter search scope for time-consuming methods (\textit{e.g.} CoTTA, MEMO) to optimize efficiency and ensure fairness. For the crucial hyperparameter batch size of TTBA and OTTA methods, we fixed it to 64 (except for TTIA method MEMO and PredBN+). Since OTTA is an online update paradigm, the different order of data entry will also affect the test results, so we fixed the loading order of the dataset.

\textbf{Extra Variants of TTA Algorithm}. Apart from directly evaluating the algorithms in Table \ref{tab:algorithms}, we also assess some variants of them. In order to have a clearer understanding of the current gap between OTTA and TTBA methods, we implement simple variants of OTTA by running them for multiple epochs (10 epochs in our benchmark). We use Tent-E10, T3A-E10, CoTTA-E10, EATA-E10 and SAR-E10 to represent them. In addition, while conducting the ViT-based experiments, we update LN modules instead of BN for Tent, EATA, SAR. Moreover, we set the batch size to 64 for PredBN+, which is referred to as PredBN+\dag \enspace in our benchmark.

\begin{table}[ht]
\centering
\small
\caption{The results of corruption datasets (\textbf{CIFAR-10-C}, \textbf{CIFAR-100-C}, \textbf{ImageNet-C}). \textit{Acc} and \textit{Rank} represent the average classification accuracies (\%) under 15 corruptions  and the ranking of the algorithm on this dataset, respectively. All results are obtained with the CNN-based backbone. PredBN+$\dag$ represents the PredBN+ with a batch size of 64.}
\label{table:corruption}
\resizebox{0.90\textwidth}{!}{
\begin{tabular}{llcrcrcrcr}
\toprule
\multicolumn{2}{c}{\multirow{2}{*}{Method}} & \multicolumn{2}{c}{\textbf{CIFAR-10-C}} & \multicolumn{2}{c}{\textbf{CIFAR-100-C}} & \multicolumn{2}{c}{\textbf{ImageNet-C}} & \multicolumn{2}{c}{Avg}   \\
         &  &  \textit{Acc}&    \textit{Rank}&   \textit{Acc}&   \textit{Rank}&   \textit{Acc}&   \textit{Rank}&   \textit{Acc}&   \textit{Rank}          \\ \midrule
\multicolumn{2}{c}{Source Model}           &   $56.49_{\pm0.00}$         &   19       &$53.55_{\pm0.00}$
           &  17       &  $18.16_{\pm0.01}$
      & 18    &    $42.73$ &18.0   \\ \midrule
\multirow{5}{*}{TTBA}     
&  PredBN+ \cite{schneider2020improving} &  $69.71_{\pm0.00}$
         &   15       & $55.37_{\pm0.00}$
          &  16        &   $24.11_{\pm0.00}$
        &    15        &$49.73$      &  15.3        \\
&  MEMO \cite{zhang2022memo}    &      $69.00_{\pm0.09}$
     &     16     &   $59.06_{\pm0.05}$        &  15        &   $24.78_{\pm0.01}$
        &   14       &$50.95$     &    15.0      \\
&  PredBN  &   $79.04_{\pm0.00}$
        &   14       & $63.89_{\pm0.00}$
          &   14       &  $31.66_{\pm0.00}$
         &    12           &    $58.20$ & 13.3      \\
&LAME \cite{boudiaf2022parameter}      & $56.41_{\pm0.00}$
          & 20         &   $32.58_{\pm0.00}$
        &   20       &   $17.99_{\pm0.00}$
        &    20    &$35.66$        &  20.0        \\
&  PredBN+\dag \cite{schneider2020improving}        &  $79.05_{\pm0.01}$
         &  13        &  $64.80_{\pm0.00}$
         &   11       &   $33.72_{\pm0.00}$
        &   10      &$59.19$      &  11.3        \\ \midrule
\multirow{5}{*}{OTTA}     
& Tent \cite{wang2020tent}         &   $79.78_{\pm0.13}$
        &   12       &  $68.31_{\pm0.02}$
         &    7      &    $43.80_{\pm0.01}$
       &      8   &$63.96$        &    9.0      \\
& T3A \cite{iwasawa2021test}        &   $61.17_{\pm0.00}$
        &   17       &  $37.74_{\pm0.01}$
             &  19        &    $18.08_{\pm0.01}$
       &    19      &       $39.00$&
18.3      \\
& CoTTA \cite{wang2022continual}      &   $80.67_{\pm0.08}$
        &   11       &   $64.37_{\pm0.06}$
        &    12      &   $38.93_{\pm0.17}$
        &     9        & $61.32$&
10.7      \\
& EATA \cite{niu2022efficient}       &  $81.51_{\pm0.08}$
         &  9        &  $68.71_{\pm0.02}$
         &  5        &   $48.31_{\pm0.08}$
        &   3             &  $66.18$&
5.7     \\
& SAR \cite{niu2023towards}        &  $81.99_{\pm0.04}$
         &  7        &  $68.36_{\pm0.01}$
         &   6       &   $45.90_{\pm0.66}$
        &    5            & $65.42$&
6.0       \\ \midrule
\multirow{8}{*}{TTDA}      
& SHOT \cite{liang2020we}        &    $80.84_{\pm0.00}$
       &    10      &  $69.93_{\pm0.01}$
         &   2       &  $45.08_{\pm0.18}$
         &   6            &  $65.28$&
6.0      \\
& NRC \cite{yang2021exploiting}        &   $82.05_{\pm0.02}$
        &   6       &  $66.83_{\pm0.15}$
         &  10        &   $32.12_{\pm0.04}$
        &    11             &    $60.33$&
9.0      \\
& AdaContrast \cite{chen2022contrastive}       &   $84.23_{\pm0.05}$
        &   1       &   $67.52_{\pm0.07}$
        &   9       &   $27.49_{\pm0.13}$
        &   13             &   $59.75$&
7.7     \\
& PLUE \cite{litrico2023guiding}       &   $82.32_{\pm1.55}$
        &   5       &  $63.89_{\pm0.02}$
         &  13        & $18.88_{\pm1.51}$
          &  17              &   $55.03$&
11.7       \\
& Tent-E10 \cite{wang2020tent}       &  $81.70_{\pm0.02}$
         &   8       &  $69.10_{\pm0.02}$
         &    4      & $44.90_{\pm0.02}$
          &    7            &   $65.23$&
6.3      \\
& T3A-E10 \cite{iwasawa2021test}      &   $58.86_{\pm0.00}$
        &   18       &  $39.49_{\pm0.01}$
         &  18        & $19.41_{\pm0.00}$
          &  16            &   $39.25$&
17.3      \\
& CoTTA-E10 \cite{wang2022continual}      &   $84.08_{\pm0.03}$
        &   2       &  $68.21_{\pm0.06}$
         &   8       &  $46.61_{\pm0.02}$
         &   4             &  $66.30$&
4.7      \\
& EATA-E10 \cite{niu2022efficient}      &  $82.85_{\pm0.02}$
         &  3        &  $69.60_{\pm0.03}$
         &   3       &  $51.01_{\pm0.01}$
         &  1             &      $67.62$&
2.3    \\ 
& SAR-E10 \cite{niu2023towards}      &  $82.60_{\pm0.20}$
         &  4        &   $70.01_{\pm0.01}$
        &   1       &   $50.04_{\pm0.02}$
        &   2               &    $67.55$&
2.3      \\ 
                           \bottomrule
\end{tabular}
}
\vspace{-0.5cm}
\end{table}

\subsection{Results on Corruption Datasets}
We will discuss the empirical study of corruption datasets in the subsequent section. The corresponding results can be found in Table \ref{table:corruption}.


In the TTBA methods, even though PredBN simply converts the BN status to training during adaptation, it still serves as an effective baseline, yielding improvements of 22.55\%, 10.34\%, and 13.50\% compared to the source model on CIFAR-10-C, CIFAR-100-C, and ImageNet-C, respectively. However, due to the limited batch size, PredBN+ and MEMO only exhibit marginal improvements.

Within the OTTA paradigm, EATA and SAR, both incorporating a sample selection module, achieve the highest performance. This highlights the importance of selecting reliable samples for optimization. Interestingly, apart from T3A, the multi-epoch versions of the OTTA methods demonstrate increased accuracy compared to their standard counterparts.

Clearly, owing to the advantage of comprehensively observing information across the entire domain simultaneously, TTDA methods attain the best performance on CIFAR-10-C and CIFAR-100-C. Surprisingly, by simply adapting the OTTA method for multiple epochs, the OTTA methods (such as CoTTA-E10 and SAR-E10) can achieve comparable accuracy to the best TTDA method on CIFAR-10-C and CIFAR-100-C. However, the experimental results on the ImageNet-C dataset present a different scenario, as the TTDA methods exhibit poor performance. Remarkably, with just a single forward adaptation, EATA achieves higher accuracy than the best TTDA method (SHOT) and EATA-E10 achieves the best performance on ImageNet-C.

Another noteworthy observation is that the methods based on contrastive learning (\textit{i.e.} AdaContrast, PLUE, and NRC) are not compatible with ImageNet-C. These methods demonstrate very limited improvements, not even surpassing the performance of PredBN. Additionally, T3A and LAME are methods that maintain the feature extractor of the model in the test state during the adaptation phase, solely optimizing the relationship between the model's features and the output. As indicated in Table \ref{table:corruption}, both methods result in negative optimization on CIFAR-100-C and ImageNet-C.

Based on the average accuracy and rank analysis, several significant conclusions can be drawn. Firstly, both the EATA and SAR variants outperform other TTDA methods and attain the highest scores. Furthermore, the CoTTA model exhibits substantial improvement when trained for multiple epochs, as evidenced by its rise in average accuracy from $61.32\%$ to $66.30\%$. Conversely, T3A only shows marginal improvement under the same conditions.

\setlength{\tabcolsep}{3.0pt}
	\begin{table*}[htbp]
		\centering
		\small
		\caption{Classification accuracies (\%) on \textbf{Office-Home} dataset.}.
		\label{table:home}
		\resizebox{1.0\textwidth}{!}{
			\begin{tabular}{llccccccccccccc}
				\toprule
				\multicolumn{2}{c}{Method}   & A$\to$C & A$\to$P & A$\to$R & C$\to$A & C$\to$P & C$\to$R & P$\to$A & P$\to$C & P$\to$R & R$\to$A & R$\to$C & R$\to$P & Avg. \\
				\midrule
				\multicolumn{2}{c}{Source Model} & $46.1_{\pm1.2}$ & $65.6_{\pm0.5}$	 & $73.8_{\pm0.5}$
 & $51.8_{\pm0.0}$ & $61.5_{\pm0.0}$
 & $63.6_{\pm0.0}$
 & $52.2_{\pm0.0}$
 & $41.9_{\pm0.0}$
 & $72.7_{\pm0.0}$
 & $65.4_{\pm0.0}$
 & $45.6_{\pm0.0}$
 & $78.0_{\pm0.0}$
 & $59.8_{\pm0.2}$
 \\
                    \midrule
				\multirow{5}{1.0em}{\rotatebox{90}{TTBA}} & PredBN+ \cite{schneider2020improving} &$47.2_{\pm1.2}$
&$65.0_{\pm0.3}$
&$73.7_{\pm0.5}$
&$53.9_{\pm0.1}$&	$61.6_{\pm0.0}$&	$63.7_{\pm0.0}$&	$53.7_{\pm0.0}$	&$42.3_{\pm0.0}$&	$73.0_{\pm0.0}$&	$65.3_{\pm0.0}$&	$46.8_{\pm0.0}$	&$76.8_{\pm0.0}$&	$60.3_{\pm0.2}$
 \\
                    & MEMO \cite{zhang2022memo} &$45.7_{\pm0.6}$&	$63.4_{\pm0.7}$	&$72.2_{\pm0.4}$	&$54.1_{\pm0.3}$	&$60.5_{\pm0.4}$	&$63.1_{\pm0.0}$	&$52.7_{\pm0.3}$	&$39.0_{\pm0.1}$	&$71.6_{\pm0.4}$	&$65.5_{\pm0.4}$	&$43.2_{\pm0.3}$	&$74.8_{\pm0.2}$	&$58.8_{\pm0.1}$
 \\
                    & PredBN \cite{nado2020evaluating} &$43.3_{\pm0.4}$	&$60.4_{\pm0.3}$	&$71.5_{\pm0.5}$	&$54.1_{\pm0.7}$	&$61.1_{\pm0.5}$	&$65.1_{\pm0.7}$	&$53.8_{\pm0.7}$	&$42.2_{\pm0.3}$	&$72.1_{\pm0.5}$	&$65.2_{\pm0.4}$	&$47.7_{\pm0.8}$	&$75.7_{\pm0.3}$	&$59.3_{\pm0.2}$
\\
                    & LAME \cite{boudiaf2022parameter} &$45.9_{\pm0.3}$	&$66.0_{\pm0.5}$	&$73.58_{\pm0.6}$	&$52.2_{\pm0.2}$	&$62.1_{\pm0.2}$	&$63.9_{\pm0.5}$	&$52.5_{\pm0.3}$	&$42.0_{\pm0.4}$	&$72.6_{\pm0.2}$	&$65.1_{\pm0.3}$	&$45.3_{\pm0.1}$	&$77.6_{\pm0.1}$	&$59.8_{\pm0.1}$
\\
                    & PredBN+\dag \cite{schneider2020improving}
                    &$46.8_{\pm0.4}$	&$65.6_{\pm0.2}$	&$73.8_{\pm0.4}$	&$53.7_{\pm0.4}$	&$62.2_{\pm0.2}$	&$64.7_{\pm0.1}$	&$54.0_{\pm0.5}$	&$43.7_{\pm0.5}$	&$72.9_{\pm0.1}$	&$65.9_{\pm0.2}$	&$49.1_{\pm0.5}$	&$77.8_{\pm0.2}$	&$60.9_{\pm0.1}$
 \\
                    \midrule
                    \multirow{6}{1.0em}{\rotatebox{90}{OTTA}} & Tent \cite{wang2020tent}
                    &$47.6_{\pm1.1}$	&$63.2_{\pm0.7}$	&$72.3_{\pm0.7}$	&$57.1_{\pm0.3}$	&$63.7_{\pm0.1}$	&$65.9_{\pm0.4}$	&$55.9_{\pm0.6}$	&$46.6_{\pm0.1}$	&$72.7_{\pm0.4}$	&$67.7_{\pm0.8}$	&$51.8_{\pm0.7}$	&$77.1_{\pm0.3}$	&$61.7_{\pm0.2}$
 \\
                    & T3A \cite{iwasawa2021test} &$49.7_{\pm0.1}$	&$73.2_{\pm0.3}$	&$77.0_{\pm0.2}$	&$55.5_{\pm0.1}$	&$67.7_{\pm0.2}$	&$68.5_{\pm0.3}$	&$56.6_{\pm0.5}$	&$45.1_{\pm0.4}$	&$75.7_{\pm0.1}$	&$67.0_{\pm0.4}$	&$49.6_{\pm0.3}$	&$78.0_{\pm0.4}$	&$63.8_{\pm0.1}$
\\
                    & CoTTA \cite{wang2022continual} &$44.5_{\pm0.2}$	&$62.5_{\pm0.7}$	&$72.3_{\pm0.3}$	&$54.9_{\pm0.6}$	&$63.1_{\pm0.8}$	&$65.9_{\pm0.3}$	&$54.2_{\pm0.4}$	&$43.5_{\pm0.2}$	&$73.0_{\pm0.4}$	&$66.0_{\pm0.6}$	&$49.5_{\pm1.4}$	&$76.7_{\pm0.1}$	&$60.5_{\pm0.1}$
\\
                    & EATA \cite{niu2022efficient} &$46.4_{\pm0.8}$	&$62.6_{\pm0.4}$	&$72.2_{\pm0.2}$	&$55.3_{\pm0.4}$	&$62.9_{\pm0.3}$	&$65.8_{\pm0.8}$	&$53.8_{\pm0.9}$	&$43.4_{\pm1.3}$	&$72.5_{\pm0.2}$	&$66.5_{\pm0.6}$	&$50.5_{\pm0.2}$	&$76.4_{\pm0.2}$	&$60.7_{\pm0.1}$
\\
                    & SAR \cite{niu2023towards}&$45.3_{\pm0.7}$	&$61.9_{\pm0.2}$	&$71.9_{\pm0.3}$	&$55.4_{\pm0.5}$	&$62.9_{\pm0.3}$	&$65.7_{\pm0.3}$	&$53.7_{\pm0.3}$	&$42.7_{\pm1.2}$	&$72.1_{\pm0.7}$	&$66.4_{\pm1.0}$	&$49.3_{\pm1.5}$	&$76.2_{\pm0.2}$	&$60.3_{\pm0.2}$
 \\
				\midrule
                    \multirow{8}{1.0em}{\rotatebox{90}{TTDA}} & SHOT \cite{liang2020we}&$56.2_{\pm0.6}$	&$75.8_{\pm0.1}$	&$79.2_{\pm0.4}$	&$66.0_{\pm1.2}$	&$74.0_{\pm1.1}$	&$74.3_{\pm0.8}$	&$64.9_{\pm0.7}$	&$54.0_{\pm0.5}$	&$79.1_{\pm0.8}$	&$71.7_{\pm0.3}$	&$56.8_{\pm0.6}$	&$82.0_{\pm0.2}$	&$69.4_{\pm0.5}$
 \\
                    & NRC \cite{yang2021exploiting}& $55.4_{\pm0.3}$	&$77.3_{\pm0.3}$	&$79.1_{\pm0.7}$	&$65.1_{\pm0.7}$	&$75.1_{\pm1.0}$	&$77.1_{\pm0.5}$	&$61.9_{\pm1.7}$	&$54.7_{\pm1.1}$	&$79.3_{\pm0.2}$	&$69.3_{\pm0.8}$	&$56.8_{\pm0.1}$	&$83.8_{\pm0.3}$	&$69.6_{\pm0.2}$
\\
                    & AdaContrast \cite{chen2022contrastive} &$53.0_{\pm0.8}$	&$73.9_{\pm05}$	&$77.5_{\pm0.8}$	&$62.8_{\pm0.5}$	&$72.7_{\pm0.0}$	&$72.4_{\pm1.0}$	&$62.1_{\pm0.8}$	&$51.0_{\pm0.1}$	&$77.9_{\pm0.8}$	&$70.0_{\pm1.1}$	&$56.5_{\pm0.7}$	&$81.5_{\pm0.5}$	&$67.6_{\pm0.1}$
\\
                    & PLUE \cite{litrico2023guiding} &$45.8_{\pm1.3}$	&$70.1_{\pm0.3}$	&$75.7_{\pm0.3}$	&$63.0_{\pm0.6}$	&$71.1_{\pm2.0}$	&$71.7_{\pm0.8}$	&$61.8_{\pm0.3}$	&$44.4_{\pm0.7}$	&$75.9_{\pm0.2}$	&$66.5_{\pm1.4}$	&$48.8_{\pm1.5}$	&$77.7_{\pm0.4}$	&$64.5_{\pm0.6}$
\\
                    & Tent-E10 \cite{wang2020tent} &$49.4_{\pm0.9}$	&$64.8_{\pm0.1}$	&$72.8_{\pm0.3}$	&$57.8_{\pm0.2}$	&$65.1_{\pm0.3}$	&$66.1_{\pm0.3}$	&$56.3_{\pm0.4}$	&$48.5_{\pm0.1}$	&$73.4_{\pm0.3}$	&$68.2_{\pm0.6}$	&$53.3_{\pm1.0}$	&$77.7_{\pm0.4}$	&$62.8_{\pm0.1}$
\\
                    & T3A-E10 \cite{iwasawa2021test}&$50.3_{\pm0.0}$	&$74.7_{\pm0.0}$	&$77.8_{\pm0.1}$	&$57.6_{\pm0.1}$	&$69.2_{\pm0.2}$	&$69.3_{\pm0.2}$	&$59.0_{\pm0.3}$	&$46.1_{\pm0.1}$	&$76.0_{\pm0.0}$	&$68.1_{\pm0.2}$	&$50.0_{\pm0.1}$	&$80.0_{\pm0.1}$	&$64.8_{\pm0.0}$
 \\
                    & CoTTA-E10 \cite{wang2022continual} &$46.0_{\pm0.5}$	&$63.2_{\pm0.3}$	&$72.9_{\pm0.2}$	&$55.2_{\pm0.4}$	&$64.0_{\pm0.3}$	&$66.7_{\pm0.3}$	&$54.9_{\pm0.5}$	&$45.0_{\pm0.4}$	&$74.2_{\pm0.5}$	&$66.5_{\pm0.5}$	&$50.0_{\pm1.0}$	&$77.8_{\pm0.4}$	&$61.3_{\pm0.1}$
\\
                    & EATA-E10 \cite{niu2022efficient} &$49.3_{\pm0.7}$	&$63.5_{\pm1.4}$	&$71.1_{\pm0.8}$	&$55.7_{\pm0.7}$	&$62.8_{\pm1.4}$	&$64.0_{\pm0.7}$	&$55.2_{\pm1.7}$	&$47.1_{\pm0.9}$	&$70.8_{\pm0.4}$	&$67.8_{\pm1.0}$	&$52.0_{\pm1.1}$	&$76.0_{\pm0.3}$	&$61.2_{\pm0.2}$
\\
                    & SAR-E10 \cite{niu2023towards} &$46.3_{\pm0.2}$	&$61.4_{\pm1.0}$	&$70.2_{\pm0.8}$	&$55.4_{\pm0.5}$	&$61.7_{\pm0.3}$	&$64.4_{\pm0.2}$	&$53.0_{\pm1.2}$	&$42.8_{\pm1.7}$	&$70.4_{\pm0.7}$	&$66.2_{\pm0.6}$	&$51.0_{\pm1.9}$	&$75.4_{\pm0.3}$	&$59.9_{\pm0.2}$
\\                    
				\bottomrule
			\end{tabular}
}
		\vspace{-5pt}
	\end{table*}

\setlength{\tabcolsep}{3.0pt}
	\begin{table*}[htbp]
		\centering
		\small
		\caption{Classification accuracies (\%) on \textbf{DomainNet126} dataset.}
		\label{table:domainnet}
		\resizebox{1.0\textwidth}{!}{
			\begin{tabular}{llccccccccccccc}
				\toprule
				\multicolumn{2}{c}{Method}   & C$\to$P & C$\to$R & C$\to$S & P$\to$C & P$\to$R & P$\to$S & R$\to$C & R$\to$P & R$\to$S & S$\to$C & S$\to$P & S$\to$R & Avg. \\
				\midrule
				\multicolumn{2}{c}{Source Model} &$49.1_{\pm0.0}$&	$62.0_{\pm0.0}$	&$50.5_{\pm0.0}$&	$56.7_{\pm0.0}$	&$74.9_{\pm0.0}$&	$48.2_{\pm0.0}$	&$58.3_{\pm0.0}$&	$53.0_{\pm0.0}$	&$60.0_{\pm0.0}$&	$57.2_{\pm0.0}$	&$62.7_{\pm0.0}$&	$47.8_{\pm0.0}$	&$56.7_{\pm0.0}$
 
 \\
                    \midrule
				\multirow{5}{1.0em}{\rotatebox{90}{TTBA}} & PredBN+ \cite{schneider2020improving} &$49.6_{\pm0.0}$&	$62.1_{\pm0.0}$	&$50.4_{\pm0.0}$&	$57.4_{\pm0.0}$	&$74.6_{\pm0.0}$&	$51.4_{\pm0.0}$	&$59.5_{\pm0.0}$&	$54.3_{\pm0.0}$	&$61.1_{\pm0.0}$&	$57.5_{\pm0.0}$	&$62.7_{\pm0.0}$&	$48.8_{\pm0.0}$	&$57.5_{\pm0.0}$

 \\
                    & MEMO \cite{zhang2022memo} &$48.1_{\pm0.0}$	&$59.6_{\pm0.0}$	&$49.0_{\pm0.1}$	&$54.8_{\pm0.0}$	&$71.9_{\pm0.0}$	&$52.3_{\pm0.1}$	&$55.3_{\pm0.0}$	&$50.6_{\pm0.1}$	&$57.4_{\pm0.0}$	&$54.6_{\pm0.2}$	&$62.0_{\pm0.0}$	&$48.0_{\pm0.0}$	&$55.3_{\pm0.0}$

 \\
                    & PredBN \cite{nado2020evaluating} &$51.8_{\pm0.0}$	&$64.7_{\pm0.0}$	&$51.5_{\pm0.0}$	&$55.4_{\pm0.0}$	&$73.7_{\pm0.0}$	&$54.5_{\pm0.0}$	&$62.0_{\pm0.0}$	&$61.7_{\pm0.0}$	&$67.5_{\pm0.0}$	&$54.7_{\pm0.0}$	&$62.8_{\pm0.0}$	&$47.6_{\pm0.0}$	&$59.0_{\pm0.0}$

\\
                    & LAME \cite{boudiaf2022parameter} &$48.7_{\pm0.0}$	&$61.4_{\pm0.0}$	&$49.9_{\pm0.0}$	&$56.4_{\pm0.0}$	&$74.5_{\pm0.0}$	&$46.5_{\pm0.0}$	&$58.0_{\pm0.0}$	&$52.4_{\pm0.0}$	&$59.4_{\pm0.0}$	&$56.7_{\pm0.0}$	&$62.4_{\pm0.0}$	&$47.3_{\pm0.0}$	&$56.1_{\pm0.0}$

\\
                    & PredBN+\dag \cite{schneider2020improving}
                    &$52.5_{\pm0.0}$	&$65.0_{\pm0.0}$	&$53.3_{\pm0.0}$	&$58.2_{\pm0.0}$	&$75.0_{\pm0.0}$	&$57.4_{\pm0.0}$	&$63.5_{\pm0.0}$	&$61.4_{\pm0.0}$	&$67.2_{\pm0.0}$	&$57.5_{\pm0.0}$	&$64.1_{\pm0.0}$	&$50.0_{\pm0.0}$	&$60.4_{\pm0.0}$

 \\
                    \midrule
                    \multirow{6}{1.0em}{\rotatebox{90}{OTTA}} & Tent \cite{wang2020tent}
                    &$53.1_{\pm0.0}$	&$64.1_{\pm0.0}$	&$53.1_{\pm0.0}$
                    &$57.3_{\pm0.0}$	&$73.9_{\pm0.0}$	&$56.8_{\pm0.0}$	&$63.1_{\pm0.0}$	&$62.0_{\pm0.0}$	&$66.6_{\pm0.0}$	&$58.0_{\pm0.0}$	&$65.6_{\pm0.0}$	&$52.0_{\pm0.0}$	&$60.5_{\pm0.0}$

 \\
                    & T3A \cite{iwasawa2021test} &$55.4_{\pm0.0}$&	$69.2_{\pm0.0}$	&$56.1_{\pm0.0}$&	$60.5_{\pm0.0}$	&$77.7_{\pm0.0}$&	$53.6_{\pm0.0}$	&$64.1_{\pm0.0}$&	$58.8_{\pm0.0}$	&$67.3_{\pm0.0}$&	$59.1_{\pm0.0}$	&$63.8_{\pm0.0}$&	$49.7_{\pm0.0}$	&$61.3_{\pm0.0}$

\\
                    & CoTTA \cite{wang2022continual} &$56.0_{\pm0.1}$	&$68.6_{\pm0.0}$	&$54.6_{\pm0.2}$	&$59.0_{\pm0.1}$	&$74.2_{\pm0.4}$	&$57.1_{\pm0.1}$	&$64.2_{\pm0.1}$	&$64.1_{\pm0.0}$	&$69.6_{\pm0.0}$	&$55.5_{\pm0.2}$	&$59.1_{\pm0.3}$	&$49.6_{\pm0.2}$	&$61.0_{\pm0.0}$

\\
                    & EATA \cite{niu2022efficient} &$54.2_{\pm0.0}$	&$65.2_{\pm0.0}$	&$54.1_{\pm0.4}$	&$58.4_{\pm0.2}$	&$73.3_{\pm0.1}$	&$57.2_{\pm0.2}$	&$63.4_{\pm0.1}$	&$62.4_{\pm0.1}$	&$67.9_{\pm0.0}$	&$57.6_{\pm0.3}$	&$64.3_{\pm0.5}$	&$52.2_{\pm0.4}$	&$60.9_{\pm0.1}$

\\
                    & SAR \cite{niu2023towards}&$52.9_{\pm0.0}$	&$64.8_{\pm0.0}$	&$53.1_{\pm0.1}$	&$56.7_{\pm0.1}$	&$73.7_{\pm0.0}$	&$56.8_{\pm0.2}$&	$63.0_{\pm0.1}$	&$62.1_{\pm0.0}$&	$67.5_{\pm0.0}$	&$56.7_{\pm0.1}$&	$64.4_{\pm0.1}$	&$51.1_{\pm0.0}$&	$60.2_{\pm0.0}$

 \\
				\midrule
                    \multirow{8}{1.0em}{\rotatebox{90}{TTDA}} & SHOT \cite{liang2020we}&$62.4_{\pm1.1}$	&$76.2_{\pm1.5}$	&$61.4_{\pm0.2}$	&$68.2_{\pm0.0}$	&$81.2_{\pm0.5}$	&$62.0_{\pm0.9}$	&$71.4_{\pm0.3}$	&$66.1_{\pm0.0}$	&$75.8_{\pm0.7}$	&$65.6_{\pm0.4}$&	$66.8_{\pm0.2}$	&$55.5_{\pm0.4}$&	$67.7_{\pm0.1}$

 \\
                    & NRC \cite{yang2021exploiting}&$59.5_{\pm0.0}$	&$73.7_{\pm0.1}$	&$57.2_{\pm0.1}$	&$62.4_{\pm0.1}$	&$80.3_{\pm0.0}$&	$59.8_{\pm0.1}$	&$69.3_{\pm0.2}$&	$66.9_{\pm0.0}$	&$75.1_{\pm0.0}$&	$62.6_{\pm0.0}$	&$67.8_{\pm0.1}$&	$54.5_{\pm0.2}$	&$65.8_{\pm0.0}$
 
\\
                    & AdaContrast \cite{chen2022contrastive} &$59.9_{\pm0.8}$	&$72.9_{\pm0.1}$	&$60.6_{\pm0.0}$&	$70.6_{\pm0.3}$	&$78.7_{\pm0.6}$&	$66.3_{\pm0.1}$	&$73.3_{\pm0.7}$&	$66.7_{\pm0.2}$	&$75.2_{\pm0.1}$&	$72.8_{\pm0.0}$	&$70.7_{\pm1.0}$&	$62.3_{\pm0.3}$	&$69.2_{\pm0.1}$

\\
                    & PLUE \cite{litrico2023guiding} &$60.5_{\pm0.4}$&	$70.8_{\pm0.4}$	&$59.4_{\pm2.2}$&	$63.8_{\pm2.1}$	&$72.8_{\pm0.4}$&	$61.2_{\pm0.4}$	&$70.8_{\pm2.1}$&	$65.6_{\pm0.3}$	&$71.9_{\pm0.7}$&	$56.1_{\pm4.0}$	&$60.4_{\pm0.9}$&	$51.7_{\pm1.5}$	&$63.7_{\pm0.3}$

\\
                    & Tent-E10 \cite{wang2020tent} &$51.2_{\pm0.0}$&	$57.2_{\pm0.0}$	&$52.2_{\pm0.0}$&	$58.5_{\pm0.0}$	&$69.9_{\pm0.0}$&	$55.6_{\pm0.0}$	&$62.9_{\pm0.0}$&	$56.9_{\pm0.0}$	&$58.7_{\pm0.0}$&	$58.9_{\pm0.0}$	&$65.7_{\pm0.0}$&	$53.1_{\pm0.0}$	&$58.4_{\pm0.0}$

\\
                    & T3A-E10 \cite{iwasawa2021test}&$54.8_{\pm0.0}$	&$67.1_{\pm0.0}$&	$55.6_{\pm0.0}$	&$60.1_{\pm0.0}$&	$76.9_{\pm0.0}$	&$52.0_{\pm0.0}$&	$64.1_{\pm0.0}$	&$58.1_{\pm0.0}$&	$66.0_{\pm0.0}$	&$58.6_{\pm0.0}$&	$63.5_{\pm0.0}$	&$48.5_{\pm0.0}$&	$60.4_{\pm0.0}$

 \\
                    & CoTTA-E10 \cite{wang2022continual} &$61.7_{\pm0.1}$&	$72.3_{\pm0.0}$	&$59.3_{\pm0.2}$&	$65.4_{\pm0.7}$	&$76.8_{\pm0.4}$&	$63.8_{\pm0.2}$	&$69.3_{\pm0.2}$&	$67.3_{\pm0.2}$	&$73.1_{\pm0.2}$&	$64.2_{\pm0.8}$	&$64.0_{\pm0.2}$&	$58.0_{\pm1.2}$	&$66.3_{\pm0.1}$

\\
                    & EATA-E10 \cite{niu2022efficient} &$55.1_{\pm0.0}$&	$64.7_{\pm0.1}$	&$54.5_{\pm0.1}$&	$58.6_{\pm0.5}$	&$72.4_{\pm0.2}$&	$58.0_{\pm0.1}$	&$64.0_{\pm0.1}$&	$62.1_{\pm0.1}$	&$67.8_{\pm0.0}$&	$58.8_{\pm0.2}$	&$63.3_{\pm0.4}$&	$53.5_{\pm0.7}$	&$61.1_{\pm0.0}$

\\
                    & SAR-E10 \cite{niu2023towards} &$53.0_{\pm0.0}$&	$62.8_{\pm0.0}$	&$53.1_{\pm0.0}$&	$57.4_{\pm0.5}$	&$73.8_{\pm0.0}$&	$55.8_{\pm0.7}$	&$63.7_{\pm0.0}$&	$62.2_{\pm0.0}$	&$67.3_{\pm0.2}$&	$57.2_{\pm0.2}$	&$64.4_{\pm0.4}$&	$50.4_{\pm0.0}$	&$60.1_{\pm0.1}$   
                    \\
				\bottomrule
			\end{tabular}
}
		\vspace{-5pt}
	\end{table*}
\subsection{Results on Natural Shift Datasets}
To demonstrate the impact of shift type on the algorithm, we conducted experiments on two natural shift datasets: Office-Home and DomainNet126. The experimental results are presented in Table \ref{table:home} and Table \ref{table:domainnet}. A noticeable disparity can be observed between the results obtained from the natural-shift datasets and the corruption datasets. Consequently, we will now address these discrepancies individually.

\textbf{PredBN exhibits inferior performance compared to its performance on corruption datasets}. In the case of DomainNet126, PredBN only yields a modest accuracy increase of $2.3\%$, which is considerably limited compared to its performance on the corruption datasets. Furthermore, PredBN experiences a decay in its effectiveness on Office-Home, with even poorer results than the source-only approach. This outcome may be attributed to the contrasting nature of corruption datasets and natural shift datasets. Compared with the Corruption dataset, the distribution shift of the natural-shift dataset is reflected in the mean and variance changes on the BN layer are not significant.


\textbf{T3A has emerged as the most effective approach within OTTA paradigms}. While its performance on corruption datasets falls behind that of other OTTA algorithms, it outshines them when it comes to natural-shift datasets. This superiority can be attributed to several reasons. Firstly, unlike other OTTA algorithms that rely on PredBN, their performance suffers when PredBN's performance deteriorates. On the other hand, a bottleneck layer was incorporated into the backbone network, resulting in more informative feature generation. This enhancement enables a more effective optimization from feature extraction to output which T3A adopts.

\textbf{TTDA methods consistently demonstrate superior performance}. While the variant of OTTA algorithms excels in achieving the highest accuracy on corruption datasets, the top-performing algorithms for natural-shift datasets are TTDA algorithms (NRC for Office-Home and AdaContrast for DomainNet126). Moreover, many OTTA methods do not derive advantages from running multiple epochs (\textit{e.g.} SAR and Tent for DomainNet126). Furthermore, the inclusion of contrastive learning techniques, as seen in NRC and AdaContrast, contributes to their remarkable performance, likely also owing to the generation of informative feature representations.

In conclusion, the performance of TTA algorithms exhibits variability by the types of distribution shift.
Stated differently, there is presently no method that demonstrates strong efficacy across diverse types of distribution shifts. 
The generalization and robustness of existing TTA methods are yet to be substantiated, thus engendering a need for further investigation and inquiry in this domain.


\setlength{\tabcolsep}{5.0pt}
\begin{table}[ht]
\centering
\caption{Classification accuracies (\%) on corruption datasets (\textbf{CIFAR10-C, CIFAR-100-C, ImageNet-C}) under OTTA and CTTA settings. All results are an average of under 15 corruptions. Gains indicate the accuracy improvement of CTTA relative to OTTA. The best results are underlined.}
\label{table:continual}
\resizebox{0.95\textwidth}{!}{
\begin{tabular}{lccccccccc}
\toprule
\multirow{2}{*}{Method} & \multicolumn{2}{c}{\textbf{CIFAR-10-C}} & \multicolumn{2}{c}{\textbf{CIFAR100-C}} & \multicolumn{2}{c}{\textbf{ImageNet-C}} & \multicolumn{2}{c}{Average}& \multirow{2}{*}{Gains} \\
&     \begin{small}OTTA\end{small}      &  \begin{small}CTTA\end{small}        &   \begin{small}OTTA\end{small}        &   \begin{small}CTTA\end{small}       &     \begin{small}OTTA\end{small}      &  \begin{small}CTTA\end{small}        &   \begin{small}OTTA\end{small}        &     \begin{small}CTTA\end{small}  &  \\ \midrule
   Source Model               & $56.49$& $56.49$ & $53.55$& $53.55$ & $18.16$ & $18.16$ & $42.73$& $42.73$&$\Delta$ \\ \midrule
   Tent \cite{wang2020tent}               &  $79.78$         &  $80.90$        & $68.31$          &      $66.31$    &     $43.80$      &    $37.14$      &    $63.96$       &  $61.45 $&$-2.51$        \\
    T3A \cite{iwasawa2021test}             &  $ 61.17$        &   $58.58$       &     $37.74$      &       $37.08$   &  $18.08$         &    $15.37$     &    $39.00$       &   $37.01 $&$-1.99$       \\
    CoTTA \cite{wang2022continual}              &  $80.67$         &   $\underline{83.45}$       &  $64.37$         &    $67.89$      &  $38.93$         &   $33.61$       &  $ 61.32$        &  $61.65$&$+0.33$        \\
    EATA \cite{niu2022efficient}              &   $ 81.51$       &  $82.06$        &       $\underline{68.71}$    &      $\underline{68.16}$    &  $\underline{48.31}$         &     $\underline{47.82}$     &   $\underline{66.18}$        &  $\underline{66.01}$&$-0.17$        \\
     SAR \cite{niu2023towards}             &   $\underline{81.99}$        &   $82.02$       &     $ 68.36 $    &       $67.39$   &  $45.90$         &     $40.47$     &    $65.42$       &  $63.29$&$-2.13$    \\   \bottomrule
\end{tabular}
}
\end{table}
\subsection{Results for Continual Test-Time Adaptation}

\begin{wrapfigure}[20]{r}{0.6\textwidth}
\centering
\includegraphics[width=0.6\textwidth]{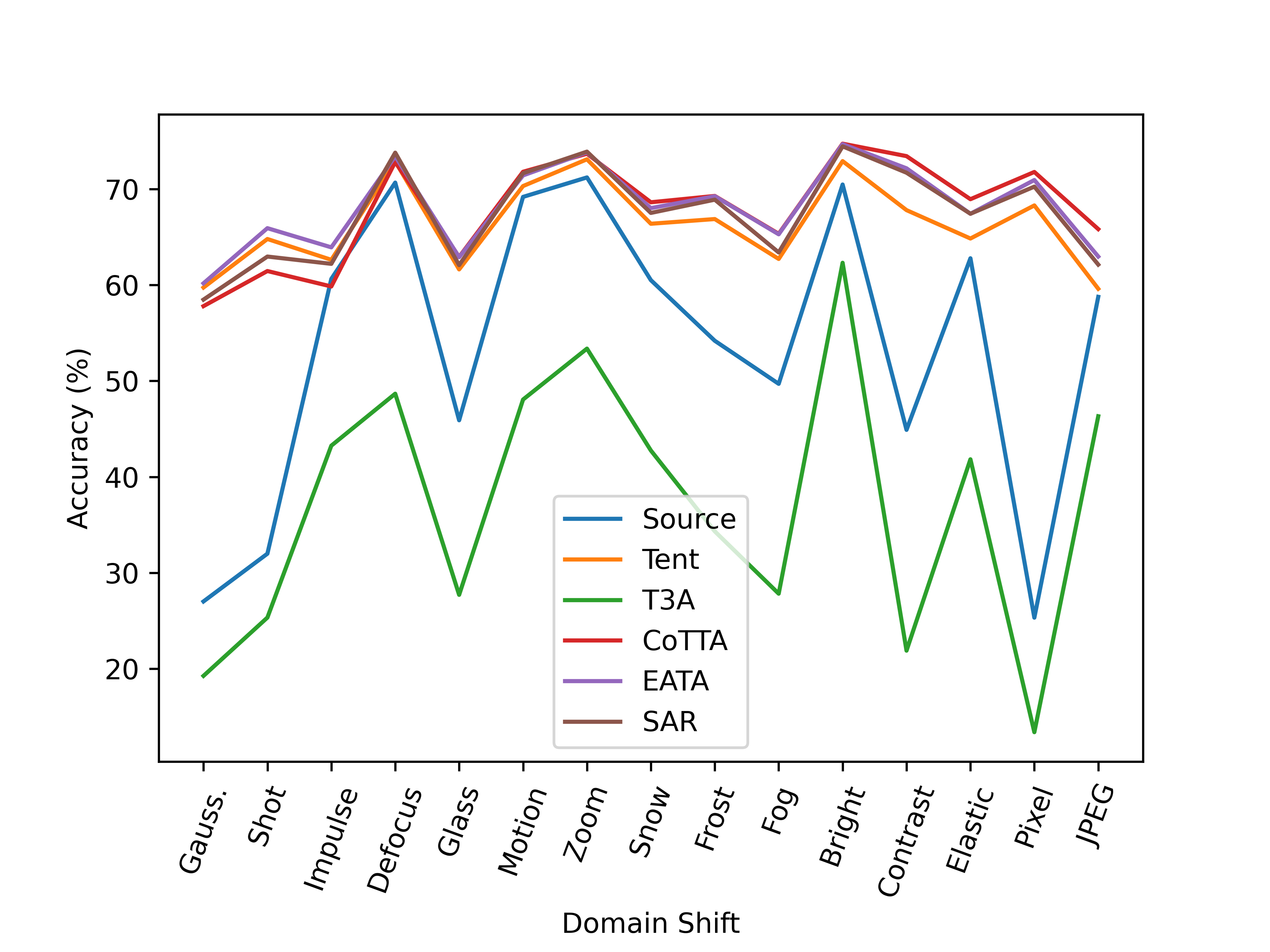}
\caption{Classification accuracies (\%) varies with domain shift on \textbf{CIFAR-100-C}.}
\label{fig:continual_cifar100}
\end{wrapfigure}

Committed to better test robustness, the target domain under the CTTA paradigm will change over time. We conduct experiments against the CTTA paradigm on corruption datasets. The experimental results are shown in Table \ref{table:continual}. Moreover, we visualize the accuracy changing for each domain on CIFAR-100-C in Figure \ref{fig:continual_cifar100}. Similar figures for CIFAR-10-C and ImageNet-C can be found in the Appendices.

The analysis of Table \ref{table:continual} reveals notable findings. Firstly, EATA achieves the highest accuracy across 4 tasks in both OTTA and CTTA, resulting in the highest average accuracy. Additionally, a comparison between OTTA and CTTA suggests that most methods exhibit lower effectiveness than OTTA or equivalent performance when operating within the CTTA framework. This can be attributed to the increased complexity and difficulty of the CTTA scenario, where models struggle to learn generalized representations based on data from different distributions in the past. However, the results obtained from CIFAR-10-C and CIFAR-100-C demonstrate that CoTTA performs better under the CTTA framework than OTTA. This indicates that CoTTA is capable of learning representations with generalization properties within the CTTA framework.

Figure \ref{fig:continual_cifar100} provides further insights into the CTTA framework. Notably, with the exception of T3A, all CTTA methods outperform the source-only model across all domains. Of particular interest is the performance of CoTTA, which exhibits moderate improvement in the early stage but surpasses other methods in the later stages. This suggests that CoTTA possesses enhanced capabilities to effectively learn from data with varying distributions.

\setlength{\tabcolsep}{3.0pt}
\begin{table}[ht]
\centering
\small
\caption{Classification accuracies (\%) on \textbf{ImageNet-C} with ViT-B/16 as backbone. The best results are underlined.}
\label{vit}
\resizebox{1.0\textwidth}{!}{
\begin{tabular}{l|m{0.5cm}m{0.5cm}m{0.5cm}m{0.5cm}m{0.5cm}m{0.5cm}m{0.5cm}m{0.5cm}m{0.5cm}m{0.5cm}m{0.5cm}m{0.5cm}m{0.5cm}m{0.5cm}m{0.5cm}|c}
\hline
Method &\rotatebox{70}{Gaussian}  &\rotatebox{70}{shot}  & \rotatebox{70}{impulse} &\rotatebox{70}{defocus}  & \rotatebox{70}{glass} & \rotatebox{70}{rotion} & \rotatebox{70}{zoom} &\rotatebox{70}{snow}  & \rotatebox{70}{frost} & \rotatebox{70}{fog} & \rotatebox{70}{brightness} & \rotatebox{70}{contrast} & \rotatebox{70}{elastic} &\rotatebox{70}{pixelate}  &\rotatebox{70}{jpeg}  &Avg.  \\ \hline
Source Model &$16.8$	&$12.0$	&$16.5$	&$29.2$	&$23.6$	&$34.0$	&$27.3$	&$15.8$	&$26.6$	&$47.5$	&$55.4$	&$44.3$	&$31.0$	&$45.1$	&$49.1$	&$31.6$
 \\ \hline
LAME \cite{boudiaf2022parameter} &$16.7$	&$12.0$	&$16.4$	&$29.0$	&$23.5$	&$33.8$	&$27.1$	&$15.7$	&$26.4$	&$47.2$	&$55.1$	&$44.1$	&$30.7$	&$45.0$	&$48.9$	&$31.4$
 \\ \hline
Tent \cite{wang2020tent} &$44.4$	&$42.8$	&$45.3$	&$52.2$	&$47.8$	&$55.5$	&$50.2$	&$18.7$	&$20.0$	&$66.5$	&$74.9$	&$64.6$	&$52.9$	&$66.8$	&$64.4$	&$51.1$
 \\
 T3A \cite{iwasawa2021test} &$16.6$	&$11.8$	&$16.4$	&$29.9$	&$24.3$	&$34.5$	&$28.5$	&$15.9$	&$27.0$	&$49.1$	&$56.1$	&$44.8$	&$33.3$	&$45.1$	&$49.4$	&$32.2$
  \\
CoTTA \cite{wang2022continual} &$40.3$	&$31.8$	&$39.6$	&$35.5$	&$33.1$	&$46.9$	&$37.3$	&$2.9$	&$46.4$	&$59.1$	&$71.7$	&$55.5$	&$46.4$	&$59.4$	&$59.0$	&$44.4$
  \\
 EATA \cite{niu2022efficient} &$54.8$	&$55.3$	&$55.6$	&$58.0$	&$59.1$	&$63.4$	&$61.5$	&$67.7$	&$66.2$	&$73.2$	&$77.9$	&$68.0$	&$68.4$	&$73.1$	&$70.3$	&$64.8$
  \\
 SAR \cite{niu2023towards} &$55.3$	&$55.7$	&$56.0$	&$58.5$	&$59.4$	&$63.6$	&$61.4$	&$67.3$	&$40.1$	&$73.3$	&$78.2$	&$68.1$	&$68.4$	&$73.4$	&$70.5$	&$63.3$
 \\ \hline
 SHOT \cite{liang2020we} &$49.8$	&$50.8$	&$51.5$	&$59.1$	&$60.1$	&$63.6$	&$63.4$	&$68.2$	&$65.8$	&$72.5$	&$78.3$	&$69.0$	&$69.2$	&$73.8$	&$70.9$	&$64.4$
  \\
 NRC \cite{yang2021exploiting} &$4.1$	&$4.1$	&$3.4$	&$15.9$	&$10.2$	&$15.5$	&$16.6$	&$5.2$	&$9.2$	&$23.1$	&$15.3$	&$40.8$	&$18.6$	&$10.3$	&$11.1$	&$13.6$
  \\
AdaContrast \cite{chen2022contrastive} &$27.1$	&$27.3$	&$28.4$	&$25.4$	&$22.2$	&$35.3$	&$28.5$	&$31.9$	&$29.6$	&$45.0$	&$54.3$	&$41.8$	&$37.5$	&$48.0$	&$42.7$	&$34.9$
  \\
 PLUE \cite{litrico2023guiding} &$42.3$	&$41.7$	&$42.2$	&$30.8$	&$27.0$	&$38.4$	&$30.3$	&$36.6$	&$40.6$	&$45.3$	&$63.8$	&$50.5$	&$41.7$	&$58.1$	&$56.1$	&$42.9$
  \\
Tent-E10 \cite{wang2020tent} &$50.6$	&$49.8$	&$51.4$	&$56.8$	&$55.2$	&$60.2$	&$55.3$	&$0.9$	&$0.5$	&$71.6$	&$77.2$	&$67.6$	&$61.6$	&$70.9$	&$68.4$	&$53.2$
  \\
T3A-E10 \cite{iwasawa2021test} &$16.9$	&$12.0$	&$16.8$	&$30.2$	&$24.9$	&$34.3$	&$28.7$	&$16.5$	&$27.4$	&$48.9$	&$55.0$	&$44.1$	&$34.8$	&$44.1$	&$49.0$	&$32.2$
 \\
EATA-E10 \cite{niu2022efficient} &$56.9$	&$58.2$	&$58.1$	&$60.0$	&$61.4$	&$66.0$	&$65.7$	&$70.3$	&$\underline{68.3}$	&$74.8$	&$78.6$	&$69.8$	&$71.5$	&$74.8$	&$71.9$	&$\underline{67.1}$
 \\
SAR-E10 \cite{niu2023towards} &$\underline{58.0}$	&$\underline{59.4}$	&$\underline{59.6}$	&$\underline{61.0}$	&$\underline{62.1}$	&$\underline{67.5}$	&$\underline{67.9}$	&$\underline{70.9}$	&$38.5$	&$\underline{75.1}$	&$\underline{78.7}$	&$\underline{70.0}$	&$\underline{72.5}$	&$\underline{75.2}$	&$\underline{72.2}$	&$65.9$
 \\ 
 \hline
\end{tabular}
}
\end{table}
\subsection{Results under Vision Transformer}
In order to showcase the profound influence of the backbone, we have conducted experiments on the ImageNet-C dataset, utilizing Vision Transformer as our chosen backbone. The outcomes of these experiments are presented in Table \ref{vit} for reference and analysis.

When utilizing ViT as the backbone, the algorithms exhibit a higher likelihood of encountering model crashes compared to ResNet-50. This is particularly evident in tasks involving snow and frost, as exemplified by Tent on snow and frost, as well as CoTTA on snow. While SAR-E10 demonstrates superior performance across most tasks, it notably underperforms in snow tasks, resulting in a lower average accuracy compared to EATA-E10.

Based on the exceptional performance exhibited by EATA and SAR, we assert that the entropy-based sample filtering module is highly effective. Additionally, we observe that achieving favorable results can be accomplished solely by updating the parameters of the LN.

\section{Conclusion and Limitations}
We construct a comprehensive benchmark for TTA, aiming to evaluate the effectiveness of the TTA method across different paradigms in the context of both corruption and natural image classification datasets. Through empirical studies, we provide valuable insights into TTA research. However, it is important to acknowledge certain limitations in our work.

\textbf{Limited downstream tasks}. Despite the existence of numerous TTA approaches proposed for various downstream tasks, such as semantic segmentation \cite{wang2021source, shin2022mm} and natural language processing \cite{singh2022addressing}, our benchmark solely focuses on the evaluation of image classification datasets.

\textbf{Lack of theoretical justification}. The conclusions drawn from our benchmark are solely based on empirical observations and do not incorporate theoretical analyses. To address this limitation, it is crucial to undertake further research that includes theoretical analysis to resolve certain issues. For instance, understanding the reasons behind the ineffectiveness of contrastive learning on ImageNet-C datasets would require a more rigorous theoretical investigation.

By acknowledging these limitations, we emphasize the need for future research to expand the scope of downstream tasks considered in TTA evaluations and incorporate theoretical analysis to enhance our understanding of the observed phenomena.

{
\small
\bibliographystyle{plain}
\bibliography{reference}

\begin{thebibliography}{10}

\bibitem{alfarra2023revisiting}
Motasem Alfarra, Hani Itani, Alejandro Pardo, Shyma Alhuwaider, Merey
  Ramazanova, Juan~C. Pérez, Zhipeng Cai, Matthias Müller, and Bernard
  Ghanem.
\newblock Revisiting test time adaptation under online evaluation, 2023.

\bibitem{boudiaf2022parameter}
Malik Boudiaf, Romain Mueller, Ismail Ben~Ayed, and Luca Bertinetto.
\newblock Parameter-free online test-time adaptation.
\newblock In {\em Proc. CVPR}, 2022.

\bibitem{chen2022contrastive}
Dian Chen, Dequan Wang, Trevor Darrell, and Sayna Ebrahimi.
\newblock Contrastive test-time adaptation.
\newblock In {\em Proc. CVPR}, 2022.

\bibitem{chen2023improved}
Liang Chen, Yong Zhang, Yibing Song, Ying Shan, and Lingqiao Liu.
\newblock Improved test-time adaptation for domain generalization.
\newblock In {\em Proc. CVPR}, 2023.

\bibitem{croce2021robustbench}
Francesco Croce, Maksym Andriushchenko, Vikash Sehwag, Edoardo Debenedetti,
  Nicolas Flammarion, Mung Chiang, Prateek Mittal, and Matthias Hein.
\newblock Robustbench: a standardized adversarial robustness benchmark.
\newblock In {\em Proc. NeurIPS}, 2021.

\bibitem{deng2009imagenet}
Jia Deng, Wei Dong, Richard Socher, Li-Jia Li, Kai Li, and Li~Fei-Fei.
\newblock Imagenet: A large-scale hierarchical image database.
\newblock In {\em Proc. CVPR}, 2009.

\bibitem{devlin2018bert}
Jacob Devlin, Ming-Wei Chang, Kenton Lee, and Kristina Toutanova.
\newblock Bert: Pre-training of deep bidirectional transformers for language
  understanding.
\newblock In {\em Proc. NAACL}, 2019.

\bibitem{ding2022proxymix}
Yuhe Ding, Lijun Sheng, Jian Liang, Aihua Zheng, and Ran He.
\newblock Proxymix: Proxy-based mixup training with label refinery for
  source-free domain adaptation.
\newblock {\em arXiv preprint arXiv:2205.14566}, 2022.

\bibitem{dosovitskiy2020image}
Alexey Dosovitskiy, Lucas Beyer, Alexander Kolesnikov, Dirk Weissenborn,
  Xiaohua Zhai, Thomas Unterthiner, Mostafa Dehghani, Matthias Minderer, Georg
  Heigold, Sylvain Gelly, et~al.
\newblock An image is worth 16x16 words: Transformers for image recognition at
  scale.
\newblock In {\em Proc. ICLR}, 2021.

\bibitem{foret2020sharpness}
Pierre Foret, Ariel Kleiner, Hossein Mobahi, and Behnam Neyshabur.
\newblock Sharpness-aware minimization for efficiently improving
  generalization.
\newblock In {\em Proc. ICLR}, 2021.

\bibitem{gan2022decorate}
Yulu Gan, Xianzheng Ma, Yihang Lou, Yan Bai, Renrui Zhang, Nian Shi, and Lin
  Luo.
\newblock Decorate the newcomers: Visual domain prompt for continual test time
  adaptation.
\newblock In {\em Proc. AAAI}, 2023.

\bibitem{ghifary2016deep}
Muhammad Ghifary, W~Bastiaan Kleijn, Mengjie Zhang, David Balduzzi, and Wen Li.
\newblock Deep reconstruction-classification networks for unsupervised domain
  adaptation.
\newblock In {\em Proc. ECCV}, 2016.

\bibitem{grandvalet2004semi}
Yves Grandvalet and Yoshua Bengio.
\newblock Semi-supervised learning by entropy minimization.
\newblock {\em Proc. NeurIPS}, 2004.

\bibitem{he2020momentum}
Kaiming He, Haoqi Fan, Yuxin Wu, Saining Xie, and Ross Girshick.
\newblock Momentum contrast for unsupervised visual representation learning.
\newblock In {\em Proc. CVPR}, 2020.

\bibitem{resnet}
Kaiming He, Xiangyu Zhang, Shaoqing Ren, and Jian Sun.
\newblock Deep residual learning for image recognition.
\newblock In {\em Proc. CVPR}, 2016.

\bibitem{hendrycks2021many}
Dan Hendrycks, Steven Basart, Norman Mu, Saurav Kadavath, Frank Wang, Evan
  Dorundo, Rahul Desai, Tyler Zhu, Samyak Parajuli, Mike Guo, et~al.
\newblock The many faces of robustness: A critical analysis of
  out-of-distribution generalization.
\newblock In {\em Proc. ICCV}, 2021.

\bibitem{hendrycks2019benchmarking}
Dan Hendrycks and Thomas Dietterich.
\newblock Benchmarking neural network robustness to common corruptions and
  perturbations.
\newblock In {\em Proc. ICLR}, 2019.

\bibitem{hinton2015distilling}
Geoffrey Hinton, Oriol Vinyals, and Jeff Dean.
\newblock Distilling the knowledge in a neural network.
\newblock {\em arXiv preprint arXiv:1503.02531}, 2015.

\bibitem{ioffe2015batch}
Sergey Ioffe and Christian Szegedy.
\newblock Batch normalization: Accelerating deep network training by reducing
  internal covariate shift.
\newblock In {\em Proc. ICML}, 2015.

\bibitem{iwasawa2021test}
Yusuke Iwasawa and Yutaka Matsuo.
\newblock Test-time classifier adjustment module for model-agnostic domain
  generalization.
\newblock In {\em Proc. NeurIPS}, 2021.

\bibitem{kar20223d}
O{\u{g}}uzhan~Fatih Kar, Teresa Yeo, Andrei Atanov, and Amir Zamir.
\newblock 3d common corruptions and data augmentation.
\newblock In {\em Proc. CVPR}, 2022.

\bibitem{krizhevsky2009learning}
A~Krizhevsky.
\newblock Learning multiple layers of features from tiny images.
\newblock {\em Master's thesis, University of Tront}, 2009.

\bibitem{lecun1989backpropagation}
Yann LeCun, Bernhard Boser, John~S Denker, Donnie Henderson, Richard~E Howard,
  Wayne Hubbard, and Lawrence~D Jackel.
\newblock Backpropagation applied to handwritten zip code recognition.
\newblock {\em Neural computation}, 1(4), 1989.

\bibitem{lee2013pseudo}
Dong-Hyun Lee et~al.
\newblock Pseudo-label: The simple and efficient semi-supervised learning
  method for deep neural networks.
\newblock In {\em Proc. ICML}, 2013.

\bibitem{lewis2019bart}
Mike Lewis, Yinhan Liu, Naman Goyal, Marjan Ghazvininejad, Abdelrahman Mohamed,
  Omer Levy, Ves Stoyanov, and Luke Zettlemoyer.
\newblock Bart: Denoising sequence-to-sequence pre-training for natural
  language generation, translation, and comprehension.
\newblock In {\em Proc. ACL}, 2020.

\bibitem{liang2023ttasurvey}
Jian Liang, Ran He, and Tieniu Tan.
\newblock A comprehensive survey on test-time adaptation under distribution
  shifts.
\newblock {\em arXiv preprint arXiv:2303.15361}, 2023.

\bibitem{liang2020we}
Jian Liang, Dapeng Hu, and Jiashi Feng.
\newblock Do we really need to access the source data? source hypothesis
  transfer for unsupervised domain adaptation.
\newblock In {\em Proc. ICML}, 2020.

\bibitem{shotplus}
Jian Liang, Dapeng Hu, Yunbo Wang, Ran He, and Jiashi Feng.
\newblock Source data-absent unsupervised domain adaptation through hypothesis
  transfer and labeling transfer.
\newblock {\em IEEE Transactions on Pattern Analysis and Machine Intelligence},
  44(11):8602--8617, 2021.

\bibitem{litrico2023guiding}
Mattia Litrico, Alessio Del~Bue, and Pietro Morerio.
\newblock Guiding pseudo-labels with uncertainty estimation for test-time
  adaptation.
\newblock In {\em Proc. CVPR}, 2023.

\bibitem{marsden2022introducing}
Robert~A Marsden, Mario D{\"o}bler, and Bin Yang.
\newblock Introducing intermediate domains for effective self-training during
  test-time.
\newblock {\em arXiv preprint arXiv:2208.07736}, 2022.

\bibitem{nado2020evaluating}
Zachary Nado, Shreyas Padhy, D~Sculley, Alexander D'Amour, Balaji
  Lakshminarayanan, and Jasper Snoek.
\newblock Evaluating prediction-time batch normalization for robustness under
  covariate shift.
\newblock In {\em Proc. ICMLW}, 2020.

\bibitem{niu2022efficient}
Shuaicheng Niu, Jiaxiang Wu, Yifan Zhang, Yaofo Chen, Shijian Zheng, Peilin
  Zhao, and Mingkui Tan.
\newblock Efficient test-time model adaptation without forgetting.
\newblock In {\em Proc. ICML}, 2022.

\bibitem{niu2023towards}
Shuaicheng Niu, Jiaxiang Wu, Yifan Zhang, Zhiquan Wen, Yaofo Chen, Peilin Zhao,
  and Mingkui Tan.
\newblock Towards stable test-time adaptation in dynamic wild world.
\newblock In {\em Proc. ICLR}, 2023.

\bibitem{peng2019moment}
Xingchao Peng, Qinxun Bai, Xide Xia, Zijun Huang, Kate Saenko, and Bo~Wang.
\newblock Moment matching for multi-source domain adaptation.
\newblock In {\em Proc. ICCV}, 2019.

\bibitem{saito2019semi}
Kuniaki Saito, Donghyun Kim, Stan Sclaroff, Trevor Darrell, and Kate Saenko.
\newblock Semi-supervised domain adaptation via minimax entropy.
\newblock In {\em Proc. ICCV}, 2019.

\bibitem{schneider2020improving}
Steffen Schneider, Evgenia Rusak, Luisa Eck, Oliver Bringmann, Wieland Brendel,
  and Matthias Bethge.
\newblock Improving robustness against common corruptions by covariate shift
  adaptation.
\newblock In {\em Proc. NeurIPS}, 2020.

\bibitem{shin2022mm}
Inkyu Shin, Yi-Hsuan Tsai, Bingbing Zhuang, Samuel Schulter, Buyu Liu, Sparsh
  Garg, In~So Kweon, and Kuk-Jin Yoon.
\newblock Mm-tta: multi-modal test-time adaptation for 3d semantic
  segmentation.
\newblock In {\em Proceedings of the IEEE/CVF Conference on Computer Vision and
  Pattern Recognition}, pages 16928--16937, 2022.

\bibitem{shorten2019survey}
Connor Shorten and Taghi~M Khoshgoftaar.
\newblock A survey on image data augmentation for deep learning.
\newblock {\em Journal of big data}, 6(1):1--48, 2019.

\bibitem{singh2022addressing}
Ayush Singh and John~E Ortega.
\newblock Addressing distribution shift at test time in pre-trained language
  models.
\newblock {\em arXiv preprint arXiv:2212.02384}, 2022.

\bibitem{strudel2021segmenter}
Robin Strudel, Ricardo Garcia, Ivan Laptev, and Cordelia Schmid.
\newblock Segmenter: Transformer for semantic segmentation.
\newblock In {\em Proc. CVPR}, 2021.

\bibitem{venkateswara2017deep}
Hemanth Venkateswara, Jose Eusebio, Shayok Chakraborty, and Sethuraman
  Panchanathan.
\newblock Deep hashing network for unsupervised domain adaptation.
\newblock In {\em Proc. CVPR}, 2017.

\bibitem{wang2020tent}
Dequan Wang, Evan Shelhamer, Shaoteng Liu, Bruno Olshausen, and Trevor Darrell.
\newblock Tent: Fully test-time adaptation by entropy minimization.
\newblock In {\em Proc. ICLR}, 2020.

\bibitem{wang2022continual}
Qin Wang, Olga Fink, Luc Van~Gool, and Dengxin Dai.
\newblock Continual test-time domain adaptation.
\newblock In {\em Proc. CVPR}, 2022.

\bibitem{wang2022out}
Xi~Wang and Laurence Aitchison.
\newblock Out of distribution robustness with pre-trained bayesian neural
  networks.
\newblock {\em arXiv preprint arXiv:2206.12361}, 2022.

\bibitem{wang2021source}
Yuxi Wang, Jian Liang, and Zhaoxiang Zhang.
\newblock Source data-free cross-domain semantic segmentation: Align, teach and
  propagate.
\newblock {\em arXiv preprint arXiv:2106.11653}, 2021.

\bibitem{xie2021segformer}
Enze Xie, Wenhai Wang, Zhiding Yu, Anima Anandkumar, Jose~M Alvarez, and Ping
  Luo.
\newblock Segformer: Simple and efficient design for semantic segmentation with
  transformers.
\newblock {\em Proc. NeurIPS}, 2021.

\bibitem{xie2020self}
Qizhe Xie, Minh-Thang Luong, Eduard Hovy, and Quoc~V Le.
\newblock Self-training with noisy student improves imagenet classification.
\newblock In {\em Proc. CVPR}, 2020.

\bibitem{xie2017aggregated}
Saining Xie, Ross Girshick, Piotr Doll{\'a}r, Zhuowen Tu, and Kaiming He.
\newblock Aggregated residual transformations for deep neural networks.
\newblock In {\em Proc. CVPR}, 2017.

\bibitem{yang2021exploiting}
Shiqi Yang, Joost van~de Weijer, Luis Herranz, Shangling Jui, et~al.
\newblock Exploiting the intrinsic neighborhood structure for source-free
  domain adaptation.
\newblock In {\em Proc. NeurIPS}, 2021.

\bibitem{zagoruyko2016wide}
Sergey Zagoruyko and Nikos Komodakis.
\newblock Wide residual networks.
\newblock {\em arXiv preprint arXiv:1605.07146}, 2016.

\bibitem{zhang2022memo}
Marvin Zhang, Sergey Levine, and Chelsea Finn.
\newblock Memo: Test time robustness via adaptation and augmentation.
\newblock In {\em Proc. NeurIPS}, 2022.

\bibitem{zhang2021adaptive}
Marvin Zhang, Henrik Marklund, Nikita Dhawan, Abhishek Gupta, Sergey Levine,
  and Chelsea Finn.
\newblock Adaptive risk minimization: Learning to adapt to domain shift.
\newblock {\em Proc. NeurIPS}, 2021.

\bibitem{zhang2015deep}
Xu~Zhang, Felix~Xinnan Yu, Shih-Fu Chang, and Shengjin Wang.
\newblock Deep transfer network: Unsupervised domain adaptation.
\newblock {\em arXiv preprint arXiv:1503.00591}, 2015.

\bibitem{zhao2023pitfalls}
Hao Zhao, Yuejiang Liu, Alexandre Alahi, and Tao Lin.
\newblock On pitfalls of test-time adaptation.
\newblock In {\em Proc. ICML}, 2023.

\end{thebibliography}
}
\newpage
\appendixpage
\begin{appendix}
    \section{Continual Test-Time Adaptation}
    \label{apdx:ctta}
\begin{figure}[htbp]
    \centering
    \includegraphics[width=0.6\textwidth]{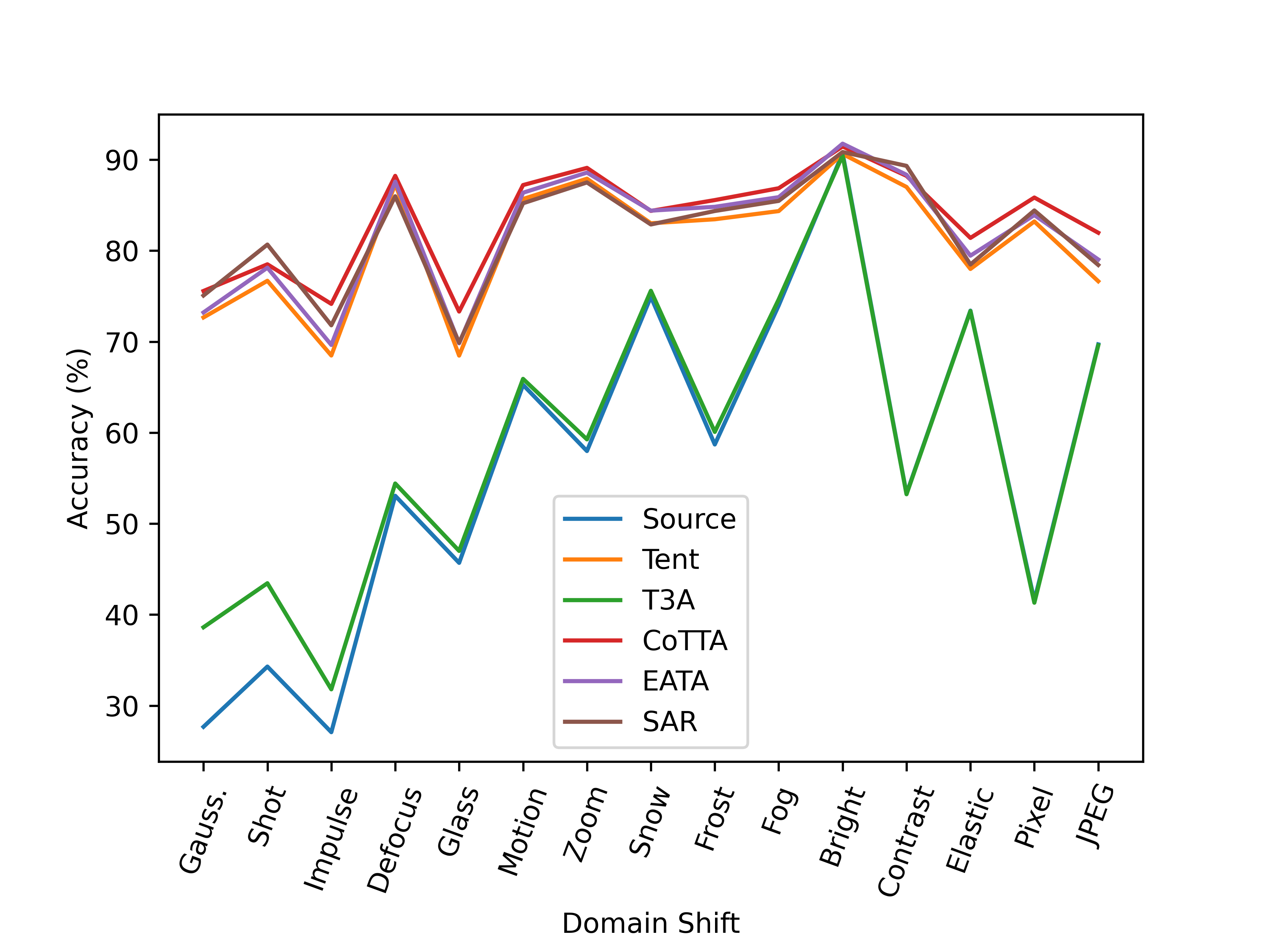}
    \caption{Classification accuracies (\%) varies with domain shift on \textbf{CIFAR-10-C}.}
    \label{fig:cifar10_c}
\end{figure}

\begin{figure}[htbp]
    \centering
    \includegraphics[width=0.6\textwidth]{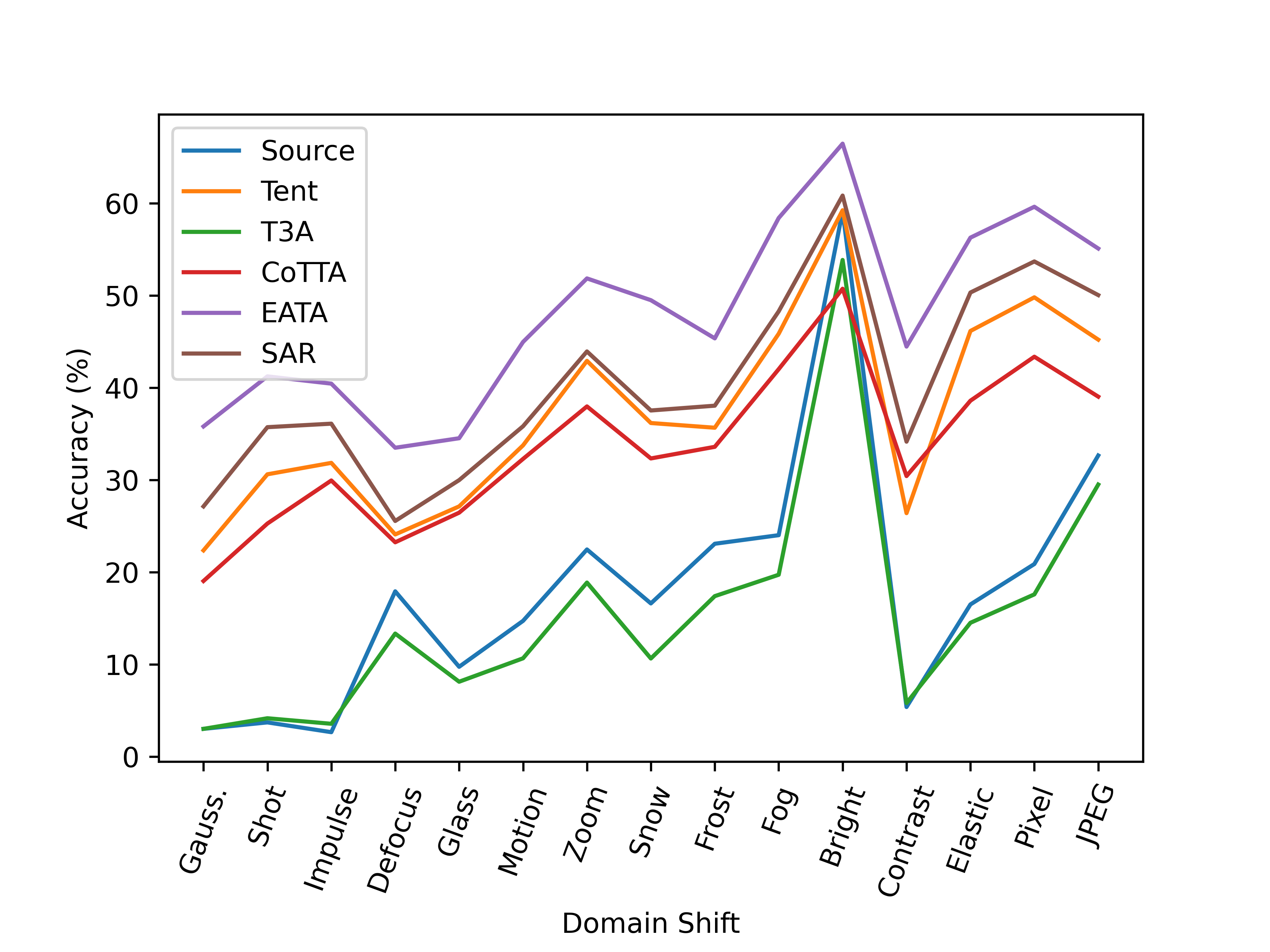}
    \caption{Classification accuracies (\%) varies with domain shift on \textbf{ImageNet-C}.}
    \label{fig:imagenet_c}
\end{figure}

The accuracy changeing for each domain on CIFAR-10-C and ImageNet-C are shown in Figure \ref{fig:cifar10_c} and Figure \ref{fig:imagenet_c}. 
We can see that all the accuracy of different algorithms have the similar trend while domain shifts. 
Algorithms except T3A have significantly improved compared to the source model. In ImageNet-C experiments, SAR showed strong performance and reached the optimum in all domains stably, while T3A declined in several domains.

\end{appendix}

\end{document}